\documentclass[10pt,twocolumn,letterpaper]{article}

\usepackage[pagenumbers]{cvpr} % To force page numbers, e.g. for an arXiv version

\usepackage{graphicx}
\usepackage{multirow}
\usepackage{amssymb}
\usepackage[normalem]{ulem}

\usepackage{tikz}
\usepackage{float}
\usepackage{algorithm}

\usepackage{listings}
\usepackage{colortbl}
\usepackage{xcolor}
\definecolor{customPink}{HTML}{FF66B2} 
\definecolor{customblue}{RGB}{34,119,188}
\definecolor{dog}{RGB}{150, 0, 200}
\definecolor{cow}{RGB}{255, 20, 147}
\definecolor{cat}{RGB}{130, 50, 60}
\definecolor{cvprblue}{rgb}{0.21,0.49,0.74}
\definecolor{custompurple}{RGB}{120, 60, 140}

\def\pz{{\phantom{0}}}
\newcommand\algcomment[1]{\def\@algcomment{\footnotesize#1}}

\newcommand{\argmin}[1]{\mathop{\arg\min}\limits_{#1}}
\newcommand{\g}[1]{\textcolor{gray!50}{#1}} % 用来表格中的灰色元素
\newcommand{\condition}{condition}
\newcommand{\conditions}{conditions}
\newcommand{\Condition}{Condition}

\usepackage[pagebackref,breaklinks,colorlinks,allcolors=cvprblue]{hyperref}

\def\paperID{1404} % *** Enter the Paper ID here
\def\confName{CVPR}
\def\confYear{2025}

\title{Decoupled Distillation to Erase: A General Unlearning Method for Any Class-centric Tasks}

\author{% 
    Yu Zhou$^{1,}$\thanks{These authors contributed equally.}\ , 
    Dian Zheng$^{1,}$\footnotemark[1]\ , 
    Qijie Mo$^{1}$, Renjie Lu$^{1}$, 
    Kun-Yu Lin$^{1,}$\thanks{These authors are corresponding authors.}\ , 
    Wei-Shi Zheng$^{1,2,3,4,}$\footnotemark[2]\ 
\\
$^1$School of Computer Science and Engineering, Sun Yat-sen University, China;\\ 
$^2$Peng Cheng Laboratory, China;\\ 
$^3$Key Laboratory of Machine Intelligence and Advanced Computing, Ministry of Education, China;\\ 
$^4$Guangdong Province Key Laboratory of Information Security Technology, China; \\ {\tt\small zhouy635@mail2.sysu.edu.cn, kunyulin14@outlook.com, wszheng@ieee.org }
}
\begin{document}

\maketitle

\begin{abstract}

In this work, we present \textbf{DE}coup\textbf{LE}d Distillation \textbf{T}o \textbf{E}rase (\textbf{DELETE}), a general and strong unlearning method for any class-centric tasks. To derive this, we first propose a theoretical framework to analyze the general form of unlearning loss and decompose it into forgetting and retention terms. Through the theoretical framework, we point out that a class of previous methods could be mainly formulated as a loss that implicitly optimizes the forgetting term while lacking supervision for the retention term, disturbing the distribution of pre-trained model and struggling to adequately preserve knowledge of the remaining classes.
To address it, we refine the retention term using ``dark knowledge” and propose a mask distillation unlearning method. By applying a mask to separate forgetting logits from retention logits, our approach optimizes both the forgetting and refined retention components simultaneously, retaining knowledge of the remaining classes while ensuring thorough forgetting of the target class.
Without access to the remaining data or intervention (\ie, used in some works), we achieve state-of-the-art performance across various benchmarks. What's more, DELETE is a general solution that can be applied to various downstream tasks, including face recognition, backdoor defense, and semantic segmentation with great performance.

\end{abstract}

\vspace{-5mm}
\section{Introduction}
\label{sec:intro}

The growing importance of ``the right to be forgotten" ~\cite{gdpr} highlights concerns around pretrained deep models, which heavily rely on user data and large-scale, unfiltered web-crawled datasets~\cite{ExcavatingAi, laion, sld}. Such reliance may compromise both user privacy and model reliability~\cite{sld,erasing,glaze}.
How to erase the influence exerted by specific subsets of training data, known as ``machine unlearning"~\cite{sisa}, has become a pressing issue. 
To circumvent the high computational expense of retraining~\cite{amnesiac}, there is a need for more efficient unlearning methods.
In this work, we focus on class-centric unlearning, which is important for applications like facial recognition, backdoor defense, and semantic segmentation, where forgetting specific classes is needed~\cite{boundary, backdoor}.

\begin{figure}[t]
    \centering
    % \rule{0.8\columnwidth}{0.45\columnwidth} % 用 \rule 来创建一个占位框
    \includegraphics[width=0.8\columnwidth]{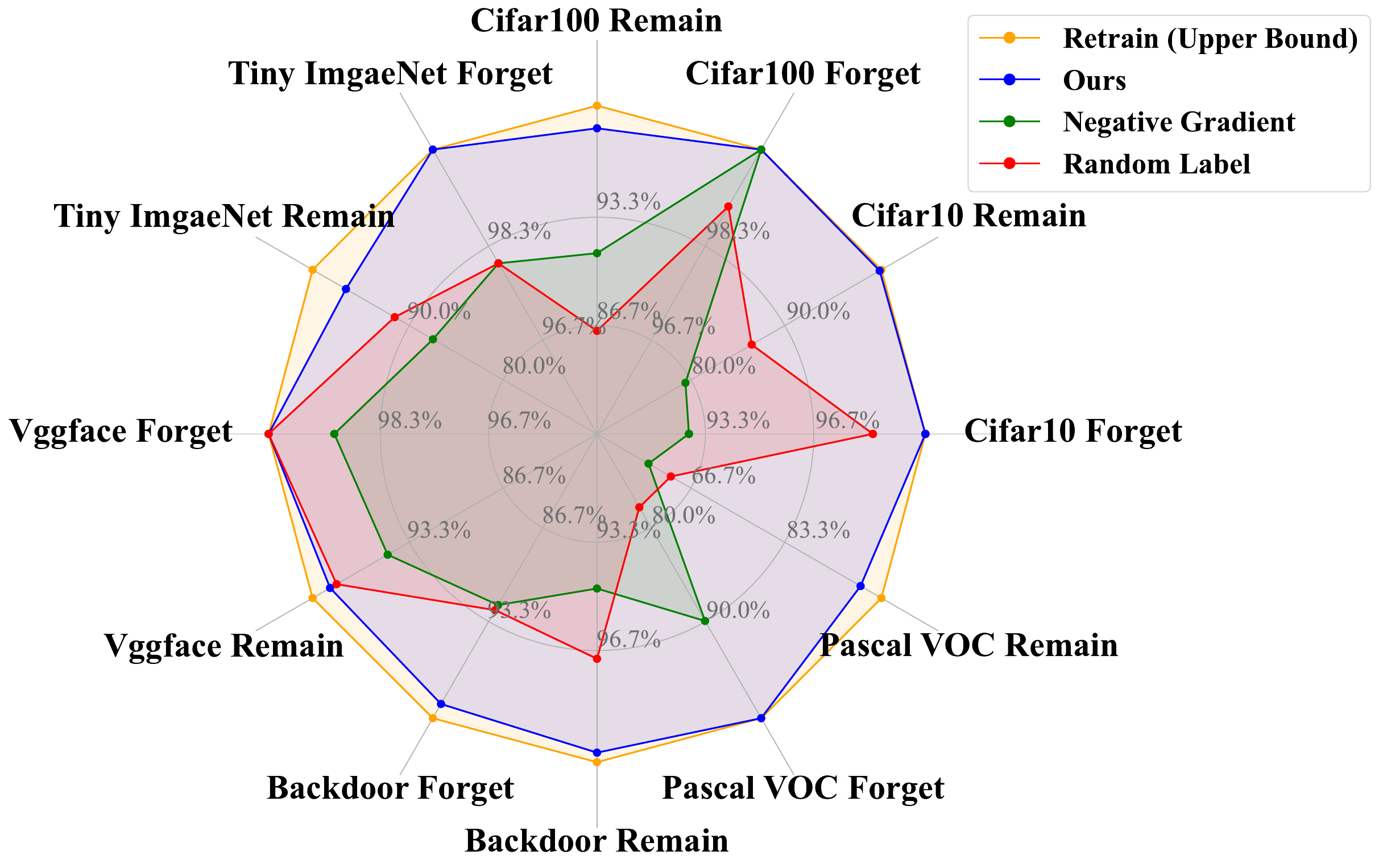}
    \vspace{-4mm}
    \caption{Performance comparison of \textcolor{blue}{our method} with previous works and \textcolor{orange}{retraining} (\ie, the \textcolor{orange}{upper bound} set to 100\%) across image classification, data poisoning, face recognition and semantic segmentation, evaluated on forgetting and remaining performance.
    }
    \label{fig:ladir}
    \vspace{-4mm}
\end{figure}

In this work, we aim to explore a general unlearning solution for any class-centric tasks, which could not be implemented by existing works as shown in Fig~\ref{fig:ladir}. To achieve it, we first propose a theoretical framework to derive the general form of unlearning loss and decouple the unlearning loss into two components: forgetting loss and retention loss, which serve as optimization objectives for forgetting and knowledge retention, respectively. Based on our theoretical framework, we reformulate previous re-label based methods~\cite{fisherforget, boundary, learn2unlearn} (\ie, a prominent category of unlearning techniques that achieve unlearning by assigning incorrect labels to forget samples.) and observe that they implicitly optimize the forgetting term but lack explicit supervision over the retention term. This limitation hinders the ability to effectively retain knowledge of remaining classes and therefore disturbs the distribution of pre-trained model~\cite{salun, boundary}. To address this, we further refine the retention loss by leveraging ``dark knowledge"~\cite{dkd}, \ie,  the relative magnitudes of the probabilities for the remaining classes.
By applying a masking strategy that separates forgetting logits from retention logits, we develop a post-hoc decoupled distillation unlearning method. 
This approach simultaneously optimizes both the forgetting and refined retention loss components, enabling effective unlearning while retaining essential knowledge. %\sout{Combining the theoretical framework and our novel unlearning loss, the DEcoupLEd distillation To Erase, termed DELETE is proposed.}
Combining the theoretical framework with our novel unlearning loss, we propose the Decoupled Distillation to Erase method, termed DELETE.

By only modifying the unlearning loss, without bells and whistles we outperform existing unlearning methods by a large margin across different benchmarks, model architectures, and forgetting settings.
Additionally, we apply our unlearning loss to downstream tasks such as face recognition, data poisoning and semantic segmentation, and achieve great performance success both in forgetting the target class and remaining others as shown in Fig~\ref{fig:ladir}, highlighting its strong generalization capabilities and promising applications in real-world scenarios. 
% To the best of our knowledge, this is the first comprehensive and fair comparison of all existing class-based machine unlearning methods conducted under the same thorough and consistent settings across a wide range of benchmarks.
% To the best of our knowledge, this is the first comprehensive and fair comparison of all existing class-centric remain-data-free machine unlearning methods conducted under the same thorough and consistent settings across a wide range of benchmarks.
To the best of our knowledge, this work presents the first comprehensive and fair comparison of all existing class-centric unlearning methods that do not rely on remaining data or intervention, evaluated under thorough settings across a wide range of benchmarks.

In summary, our main contributions are as follows:
\begin{itemize} 
    % \item Identifying the two challenges in previous work, we propose a new, stricter task setup for class-centric machine unlearning, such that no access to remaining data and pretraining phase.

    % So we further refine the retention loss using ``dark knowledge" and retaining assumption, in order to preserve knowledge without accessing the remaining data.
    % % in key observations and assumptions. 
    % Through masking to decouple the forgetting logits from the retention logits, we propose a post-hoc decoupled distillation unlearning method, which optimizes both the forgetting and refined retention loss components simultaneously. \sout{We also point out that previous methods implicitly adapted the forgetting term but lacked supervision for the retention loss.}
    
    \item We adopt a novel theoretical analysis that decomposes the unlearning loss into a forgetting term and a retention term. This inspires a new perspective on leveraging and preserving knowledge within the model based on logits. Under the analysis framework, we also point out that some previous methods implicitly adapt the forgetting term but lack supervision for the retention loss.
    
    \item Motivated by the analysis, we further refine the retention loss using ``dark knowledge" and introduce the post-hoc decoupled distillation unlearning method, termed DELETE. By masking to decouple the logits of forgotten and non-forgotten classes, we optimize both the decoupled objectives simultaneously, which leads to both effective unlearning and excellent knowledge preservation.
    
    \item Through extensive experiments across a wide range of benchmarks, we demonstrate our method's SOTA performance. Additionally, we apply it to multiple real-world downstream tasks, such as face recognition, backdoor defense, and semantic segmentation, highlighting its strong generalization capabilities.
\end{itemize}

\vspace{-1mm}
\section{Related Work}
\label{sec:related-work} 
% \subsection{Machine Unlearning}
% \vspace{-2mm}
\noindent\textbf{Machine Unlearning.}
% \notice{machine unlearn in ai safety}
% Nowadays, large pre-trained models are often trained on massive, multi-modal datasets crawled from the internet, which are typically unfiltered and may contain biased, harmful, privacy-invasive, or even poisoned data. This poses significant risks to model fairness, privacy, and potential legal violations. Machine Unlearning has emerged as a solution to eliminate the influence of harmful data from models without affecting unrelated knowledge.
Machine Unlearning is the task of removing the influence of specific data from a pretrained model~\cite{
sisa, ginart2019making, cao2015towards}. This task addresses the ``right to be forgotten”~\cite{gdpr}, which is crucial for privacy protection~\cite{sisa} and enhancing AI safety~\cite{backdoor, oesterling2024fair, gandikota2023erasing}. Retraining from scratch is a common yet naive approach, as it incurs substantial computational overhead~\cite{amnesiac, deltagrad}.
Some methods explore machine unlearning for traditional machine learning tasks, such as linear regression~\cite{baumhauer2022machine,mahadevan2021certifiable}, k-means~\cite{ginart2019making}, random forests~\cite{brophy2021machine}, and SVM~\cite{chen2019novel}. However, these methods typically leverage the convex nature of the problem or other theoretical constraints, so that can't be adopted on DNN~\cite{cao2015towards, baumhauer2022machine, zheng2025diffuvolume, zheng2024selective, lv2024spatialdreamer, liu2023generating, xu2024dexterous, wu2024economic, wu2023estimator, cao2024grids,zheng2025vbench}.

% Machine unlearning: Linear filtration for logit-based classifiers
% Machine unlearning for random forests.
% A novel online incremental and decremental learning algorithm based on variable support vector machine
% Certifiable machine unlearning for linear models

\noindent\textbf{Unlearning in DNN.}
% Consequently, there is an urgent need for fast unlearning methods that can generalize to deep learning models, which leads to the development of approximate unlearning techniques. These methods use the forget data to quickly fine-tune the model, achieving the goal of data forgetting.
%Grave记录梯度信息。Wu 提出了delta grad，使用quasi newton方法存储和减去forget data 的梯度。
% The urgent need for fast unlearning methods in DNNs lead to the development of various types of unlearning approaches. Some methods require intervention during the pre-training process. For instance, the SISA method splits the original model into multiple sub-models, training them independently. During the unlearning phase, these sub-models are selectively removed to achieve the forgetting effect. Unrolling SGD introduces additional loss terms during training, while other approaches rely on storing gradient information during pre-training, in order to facilitate and accelerate the unlearning process.
% However, these approaches increase overhead, reduce computational efficiency, and may degrade model performance, making them unsuitable for machine-as-a-service (MaaS) scenarios and  complex real-world scenarios.
The urgent need for fast unlearning methods in DNNs leads to various approaches. Some methods require intervention during the pre-training process. SISA~\cite{sisa} splits the model into multiple sub-models trained independently, and selectively removes these sub-models to unlearn. Unrolling SGD~\cite{unrollingsgd} adds extra loss terms during training, while other methods~\cite{amnesiac, deltagrad} rely on stored gradient information to facilitate and speed up unlearning. However, these designs increase computational overhead, reduce efficiency, and may degrade model performance~\cite{mu:survey, survey}, making them impractical for machine-as-a-service (MaaS) and complex real-world applications~\cite{maas1, maas2}.

To achieve effective forgetting while preserving unrelated knowledge, other methods attempt to use the remaining data. Fisher Forgetting~\cite{fisherforget} calculates the Fisher Information Matrix based on remaining data, adding noise to scrub parameters associated with forgetting data. SSD~\cite{ssd} calculates parameter importance for both forgotten and remaining data, then applies weight perturbations to unlearn. Additionally, some methods introduce regularization terms or an extra repair stage to reduce interference with unrelated knowledge~\cite{unsir,scrub}. 
However, in real-world applications, factors such as restricted data access, data expiration policies, storage costs, or privacy concerns may prevent providers from storing remaining data~\cite{survey3, learn2unlearn, economics}. In order to address these challenges, unlearning should ideally be performed without relying on such data. In these cases, the above methods become unusable or suffer from observable performance degradation~\cite{ssd, unsir, laf}.

% To achieve effective forgetting while preserving unrelated knowledge, other methods attempt to use the remaining data. Fisher Forgetting~\cite{fisherforget} calculates the Fisher Information Matrix based on remaining data, adding noise to scrub parameters associated with forgetting data. SSD~\cite{ssd} calculates parameter importance for both forgotten and remaining data, then applies weight perturbations to unlearn. Additionally, some methods introduce regularization terms or an extra repair stage to reduce interference with unrelated knowledge~\cite{unsir,scrub}. 
% However, in real-world applications, factors such as restricted data access, data expiration policies, storage costs, or privacy concerns may prevent providers from storing remaining data~\cite{survey3, learn2unlearn, economics}. In order to address these challenges, unlearning should ideally be performed without relying on such data. In these cases, the above methods become unusable or suffer from observable performance degradation~\cite{ssd, unsir, laf}.

% Thus, enabling remain data-free machine unlearning w/o pretraining intervene has become a key challenge.
Therefore, a stricter and more practical machine unlearning scenario needs to be explored.
Our method does not interfere with the pre-training process, or rely on remaining data, making it readily applicable in real-world scenarios.

\noindent\textbf{Knowledge Distillation.}
%\notice{对抗样本是深度学习攻击模型的有效方法，通过给输入图片增加人眼不可见的噪声，干扰模型的正常功能. 由于对抗样本还有分类的有效特征，被用在分类遗忘领域。}
Knowledge distillation~\cite{kd}, initially proposed by Hinton et al., is a technique that transfers knowledge from a teacher model to a student model. FitNets~\cite{fitnet} further advances this approach by incorporating intermediate features to better align teacher and student models.
% and Attention Transfer attention mechanisms
Decoupled knowledge distillation~\cite{dkd} separate the logits into TCKD 
%(Task-specific Knowledge Distillation) 
and NCKD 
%(Non-task-specific Knowledge Distillation) 
components, enhancing model capability and generalization.

% In recent years, knowledge distillation has been applied across various computer vision tasks, 
In recent years, knowledge distillation has been applied across various vision tasks,
including machine unlearning~\cite{tang2024rethinking,kdInDetect,mo2024bridge,kdInBert,kdIndiffusion,kdSurvey}. Bad Teacher~\cite{badteacher} initializes a pair of competent and incompetent teacher models, using knowledge distillation to selectively unlearn while retaining unrelated knowledge. SCRUB~\cite{scrub} leverages the teacher model’s outputs on remaining data to preserve knowledge.
Our approach decouples the unlearning loss into two distinct components
%: forgetting and retention, 
and incorporates mask to guide the distillation process. 
Unlike previous methods, we achieve selective forgetting without access to remaining data.

% For instance, the Boundary Shrink method employs adversarial attacks to generate adversarial labels, aiming to forget the target classes while preserving the knowledge of the remaining ones as much as possible. Similarly, the Learn to Unlearn method uses adversarial samples and labels as regularization terms for the remaining classes. Although these adversarial attack-based methods perform well without accessing the remaining data, they still face a trade-off between forgetting and retention, making it challenging to fully forget the target classes while maintaining strong performance on the remaining ones. 

% \section{Method}
% \label{sec:method}
% \notice{a section summary here}
\vspace{-3mm}
\section{Problem Definition for Machine Unlearning}
\vspace{-1mm}
We now define the notion of class-centric machine unlearning. Let the dataset be denoted as $\mathcal{D}=\{\mathbf x_i,  y_i\}_{i=1}^N\subseteq\mathcal{X}\times\mathcal{Y}$, consisting of $N$ data points. 
%In each data point, 
$\mathbf x\in\mathcal{X}\subset\mathbb{R}^d$ represents the input image, and $ y\in\mathcal{Y}=\{1,\cdots,K\}$ is the ground truth label of the image. Let the original classification model trained on the dataset $\mathcal{D}$ be denoted as $f_{\theta_{\mathrm o}}:\mathcal{X}\rightarrow\mathcal{Y}$, where $\theta_{\mathrm o}$ represents the model parameters.

In machine unlearning, the original dataset is divided into two non-overlapping subsets, $\mathcal{D}_{\mathrm f}$ and $\mathcal{D}_{\mathrm r}$.
%, such that $\mathcal{D}_{\mathrm f}\cap\mathcal{D}_{\mathrm r}=\emptyset$ and $\mathcal{D}_{\mathrm f}\cup\mathcal{D}_{\mathrm r}=\mathcal{D}$. 
The goal of machine unlearning is to erase the knowledge related to $\mathcal{D}_{\mathrm f}$ from the model while preserving knowledge related to $\mathcal{D}_{\mathrm r}$ remains unaffected. Some existing methods use all or part of $\mathcal{D}_{\mathrm r}$ during unlearning to
preserve the model's related knowledge. However, due to privacy concerns and  restricted data access in real-world applications~\cite{survey3, learn2unlearn, economics}, we adopt
%propose
a stricter machine unlearning setting: during unlearning, \textbf{only the data from $\mathcal{D}_{\mathrm f}$ can be used.}

As a consequence, the unlearning algorithm trains the model parameters $\theta_{\mathrm o}$ on $\mathcal{D}_{\mathrm f}$ to obtain new parameters $\theta$. The unlearned model $f_\theta$ is required to exhibit incorrect classification on $\mathcal{D}_{\mathrm f}$, while maintaining high classification accuracy on $\mathcal{D}_{\mathrm r}$.
%, as a result of the forgetting and preservation on the respective datasets. 
The model $f_{\theta^{\star}}$, retrained on $\mathcal{D}_{\mathrm r}$, is typically regarded as the upper bound for machine unlearning~\cite{boundary,necessity}. 
The output of the unlearned model $f_\theta$ should be as close as possible to this target model $f_{\theta^{\star}}$.
%\sout{in the output space.}

\vspace{-2mm}
\section{Theoretical Analysis and Method}
\vspace{-1mm}

In this section, we first present a theoretical analysis framework to explore the general form of unlearning loss and decompose it into forgetting and retention losses. Based on the theoretical analysis, we show that re-label unlearning methods are a special case of unlearning loss.
Then we suggest a forgetting \condition\  on the forgetting loss derived from experimental observations; while re-label methods also implicitly satisfy this, they lack supervision on the remaining classes. Therefore, we introduce a retention \condition\ for the retention loss. By combining both \conditions, we finally propose a mask distillation unlearning method.

% Based on this theoretical analysis, we introduce a mask-based distillation method for unlearning, where we assign approximate probability distributions to both the forgotten class and the remaining classes, in order to meet requirements of each loss term. 
% This approach demonstrates superior performance in various experiments, achieving effective forgetting of the target class while retaining knowledge of other classes. Leveraging the theoretical insights, we also observe that adversarial attack-based methods are equivalent to using one-hot vectors for supervision, which fail to provide adequate supervision for unrelated classes \( i \neq u, r \).
% \vspace{-1mm}
\subsection{Theoretical Analysis}
\vspace{-1mm}
\label{sec:theoretical analysis}
\noindent\textbf{1) Decoupling Unlearning Loss.}
In this section, inspired by DKD~\cite{dkd}, we propose the unlearning loss and decompose it into two parts: the \textbf{forgetting loss} and the \textbf{retention loss}. For a classification sample \( \mathbf{x} \in \mathcal{D}_{\mathrm f} \) from class \( u \) to be unlearned, let the target probability distribution output %by the ideal unlearned model 
be \( \mathbf{p} \in \mathbb{R}^K \), and the output of the actual unlearning model \( f_{\theta} \) be \( \mathbf{q} \in \mathbb{R}^K \). Typically, the unlearning loss should reduce 
% the gap between these distributions via 
KL divergence, given by $\mathcal{L} = \mathrm{KL}(\mathbf{p} \| \mathbf{q})$.

Let \( p_i \) denote the \( i \)-th component of the probability distribution for the target \( \mathbf{p} \).
Specifically, \( p_u \) and \( p_{\backslash u} = \sum_{i \neq u} p_i \) denote the probabilities corresponding to the forgetting class and the sum for all other classes, respectively. We denote the binary-class probabilities as \( \mathbf{p}^{(b)} = [p_u, p_{\backslash u}] \). Similarly, the predicted probabilities \( q_u \), \( q_{\backslash u} \) and the binary-class distribution \( \mathbf{q}^{(b)} = [q_u, q_{\backslash u}] \) are defined.
%follow the same notation.
For all classes \( i \neq u \), we define \( \hat{\mathbf{p}} \) and \( \hat{\mathbf{q}} \) as the re-normalized probability distributions over these \( K-1 \) classes, \ie, each element satisfies \( \hat{p}_i = \frac{p_i}{\sum_{i \neq u} p_i} \) and \( \hat{q}_i = \frac{q_i}{\sum_{i \neq u} q_i} \) .

The unlearning loss function, which reduces the gap between the model’s predicted distribution and the target distribution, can be decomposed into two parts:
\vspace{-1mm}
\begin{align}
\mathcal{L} =p_u \log\left( \frac{p_u}{q_u} \right) + \sum_{i \neq u} p_i \log\left( \frac{p_i}{q_i} \right).
\label{equ:split}
\end{align}

\vspace{-1mm} 
For the second term, leveraging the identities \( p_i = \hat{p}_i \cdot p_{\backslash u} \) and \( q_i = \hat{q}_i \cdot q_{\backslash u} \) for \( i \neq u \), we have:
\vspace{-2mm} 
\[
\begin{aligned}
\sum_{i \neq u} p_i \log\left( \frac{p_i}{q_i} \right) &= p_{\backslash u} \sum_{i \neq u} \hat{p}_i \log\left( \frac{\hat{p}_i \cdot p_{\backslash u}}{\hat{q}_i \cdot q_{\backslash u}} \right) \\
&= p_{\backslash u} \sum_{i \neq u} \hat{p}_i \left( \log\left( \frac{\hat{p}_i}{\hat{q}_i} \right) + \log\left( \frac{p_{\backslash u}}{q_{\backslash u}} \right) \right) \\
&= p_{\backslash u} \log\left( \frac{p_{\backslash u}}{q_{\backslash u}} \right) + p_{\backslash u} \sum_{i \neq u} \hat{p}_i \log\left( \frac{\hat{p}_i}{\hat{q}_i} \right).
\end{aligned}
\]

Substituting this result into the equation \cref{equ:split}, the loss function becomes:

\vspace{-5mm}
\begin{equation}
    \begin{aligned}
    \mathcal{L} &= p_u \log\left( \frac{p_u}{q_u} \right) + p_{\backslash u} \log\left( \frac{p_{\backslash u}}{q_{\backslash u}} \right) + p_{\backslash u} \sum_{i \neq u} \hat{p}_i \log\left( \frac{\hat{p}_i}{\hat{q}_{i}} \right) \\
    &= \mathrm{KL}(\mathbf p^{(b)} \| \mathbf q^{(b)}) + p_{\backslash u} \mathrm{KL}(\hat{\mathbf p} \| \hat{\mathbf q}).
    \end{aligned}
    \label{equ:decouple loss}
\end{equation}

% Thus, the original loss function is decoupled into two terms: one enforcing forgetting on the target class and another supervising the remaining classes, termed as \textbf{forgetting loss} and \textbf{retention loss}. Here, \( \mathbf{p}^{(b)} = [p_u, p_{\backslash u}] = [p_u, 1 - p_u] \) depends only on the forgotten class \( p_u \), and similarly, \( \mathbf{q}^{(b)} \) depends on \( q_u \).
Thus, we decouple the original loss function into two components: one focused on enforcing forgetting of the target class, and the other supervising the remaining classes. The former arises from the fact that \( \mathbf{p}^{(b)} = [p_u, p_{\backslash u}] = [p_u, 1 - p_u] \) depends only on the probability of the forgotten class, \( p_u \), and similarly, \( \mathbf{q}^{(b)} \) depends on \( q_u \). We refer to these components as the \textbf{forgetting loss} and \textbf{retention loss}.

\vspace{1mm}
\noindent\textbf{2) Rethinking Re-label Unlearning Method.}  In this part, we demonstrate that re-label unlearning methods, a common approach within unlearning tasks, represent a special case of the previously established theoretical framework. This allows us to leverage the conclusions drawn in \cref{equ:decouple loss}.
% Some previous 遗忘 works 尝试给$D_f$中的样本$x$重新分配一个fixed incorrect replacement class label \( r \neq u \) 标签进行学习。supervise the unlearning of \( \mathbf{x} \). 
% 例如random label 方法随机给$x$分配标签.
% some methods employ  adversarial attack methods (\eg, PGD and FGSM) for unlearning, and relabel is obtained by attacking the original sample \( \mathbf{x} \) against the frozen original model \( f_{\theta_{\mathrm o}} \).

Re-label methods assign a fixed incorrect label \( r \neq u \) to sample \( \mathbf{x} \) %in \( \mathcal{D}_f \)
, as a form of supervised unlearning. For example, random label assigns a randomly selected label to \( \mathbf{x} \)~\cite{fisherforget}. Some methods use adversarial attacks to derive replacement labels, by attacking \( \mathbf{x} \) against the frozen model \( f_{\theta_{\mathrm o}} \)~\cite{boundary, learn2unlearn}. The unlearning loss \( \mathcal{L}_{\mathrm{re-label}} \) is defined as the cross-entropy loss between \( \mathbf{x} \) and the incorrect label \( r \), formulated as:
\begin{equation}
    \begin{aligned}
    \argmin{\theta} \mathcal{L}_{\mathrm{re-label}} &= \argmin{\theta}  -\log(q_{r}).
    \label{equ:relabel loss}
    \end{aligned}
\end{equation}

\vspace{-1mm}
We then demonstrate that this is equivalent to the KL-divergence between the one-hot vector of class \( r \) and the model's output. Let the one-hot vector for class \( r \) be denoted as \( \mathbf{e}^{r} \).
%, \sout{which corresponds to the \( r \)-th column of the identity matrix \( I^{K \times K} \).} 
When the target output distribution \( \mathbf p := \mathbf e^r \), note the fact that target \( \mathbf p \) is independent of the optimization parameter \( \theta \) and the properties of one-hot vectors. In this case, we have:

\vspace{-5mm}
\[
\begin{aligned}
\argmin{\theta} \mathcal{L}_{\mathrm{re-label}} 
&= \argmin{\theta} -\log(q_r) + \log(p_r) \\
&= \argmin{\theta} \log\left(\frac{p_r}{q_r}\right) \\
% &= \argmin{\theta} p_r \log\left( \frac{p_r}{q_r} \right) \\
&= \argmin{\theta} \sum_{i=1}^K p_i \log\left( \frac{p_i}{q_i} \right).
\end{aligned}
\]

\vspace{-1mm}
By transforming the re-label unlearning loss \cref{equ:relabel loss} into a KL-divergence form as in \cref{equ:split}, we enable further discussion within the decomposition framework introduced earlier. In other words, re-label methods represent a special case of \cref{equ:decouple loss}, where the target distribution is set as a one-hot vector.

% \subsection{Implicit Assumption in Previous Method}
% \subsection{Implicit Forgetting Condition in Re-label Method}
% \label{sec:intial assumption}
\vspace{1mm}
\noindent\textbf{3) Implicit Forgetting \Condition\ in Re-labeling.}While the target distribution is not accessible during unlearning, we can approximate \( \mathbf{p} \) by observing the probability distributions generated by the retrain model, which is typically considered the gold standard for machine unlearning~\cite{boundary,necessity}.
% As shown in the results \notice{table}, after unlearning, the probability of the forgotten class \( p_u \) becomes an extremely small value approaching zero. Previous re-label methods have implicitly applied this approximation by using a one-hot target, where \( p_u \) is set as \( p_u := e_u = 0 \).
% Thus, based on the above observations, we adopt the forgetting assumption \( p_u = 0 \) on \textbf{forgetting loss} mentioned in \cref{equ:decouple loss}, as an approximation of the ideal target distribution. Consequently, we have \( p_{\backslash u} = 1 - p_u = 1 \). At this point, \( \mathbf{p}^{(b)} = [p_u, p_{\backslash u}] = [0, 1] \), denoted as \( \mathbf{e}^{(b)} = [0, 1] \). The loss function from \cref{equ:decouple loss} can then be further expressed as:
As shown in the \cref{tab:cifar10 cifar100}  in the subsequent experiments, after unlearning, the probability of the forgotten class \( p_u \) approaches {0}.
%an extremely small value approaching zero
Previous re-label methods implicitly apply this approximation by using a one-hot target, where \( p_u \) is set as \( p_u := e_u = 0 \).

Based on these observations, we formally adopt the \textbf{forgetting \condition} \( p_u = 0 \) in \textbf{forgetting loss} from \cref{equ:decouple loss}, as an approximation of target distribution. Consequently, we obtain \( p_{\backslash u} = 1 - p_u = 1 \), leading to \( \mathbf{p}^{(b)} = [p_u, p_{\backslash u}] = [0, 1] \), which we denote as \( \mathbf{e}^{(b)} = [0, 1] \). The loss function in \cref{equ:decouple loss} can be further expressed as:

\vspace{-4mm}
\begin{equation}
    \resizebox{1.05\columnwidth}{!}{$
        \begin{aligned}
        \mathcal{L}
        &= \mathrm{KL}(\mathbf e^{(b)} \| \mathbf q^{(b)}) + p_{\backslash u} \mathrm{KL}(\hat{\mathbf p} \| \hat{\mathbf q}) \\
        &= \mathrm{KL}(\mathbf e^{(b)} \| \mathbf q^{(b)}) + p_{\backslash u} \sum_{i \neq u} \hat{p}_i \log\left( \frac{\hat{p}_i}{\hat{q}_i} \right) \\
        &= \mathrm{KL}(\mathbf e^{(b)} \| \mathbf q^{(b)}) + p_{\backslash u} \hat{p}_r \log\left( \frac{\hat{p}_r}{\hat{q}_r} \right) + p_{\backslash u} \sum_{i \neq u, r} \hat{p}_i \log\left( \frac{\hat{p}_i}{\hat{q}_i} \right).
        \end{aligned}$
    }
    \label{equ:three term loss}
\end{equation}
% $$
% \mathcal L=
% \mathrm{KL}(\mathbf e^{(b)} \| \mathbf q^{(b)}) + p_{\backslash u} \hat{p}_r \log\left( \frac{\hat{p}_r}{\hat{q}_r} \right) + p_{\backslash u} \sum_{i \neq u, r} \hat{p}_i \log\left( \frac{\hat{p}_i}{\hat{q}_i} \right)
% $$
It suggests that the loss function should not only enforce forgetting the target class via the first term, but also supervise predictions for remaining classes via the second and third terms. 
This structure helps maintain the model’s predictions for other classes before and after unlearning, preventing excessive forgetting.

However, re-label unlearning methods lack this supervision for unrelated classes \( i \neq u, r \). 
Recall re-label methods are equivalent to using a one-hot vector for supervision. 
The corresponding element in the target distribution is assigned \( p_i := 0 \), (\ie, \( \hat{p}_i := 0 \)), transforming \cref{equ:three term loss} into:
\vspace{-2mm}
\[
\begin{aligned}
\mathcal{L}_{\textrm{re-label}} = \mathrm{KL}(\mathbf e^{(b)} \| \mathbf q^{(b)}) + p_{\backslash u} \hat p_r \log\left( \frac{\hat p_r}{\hat q_r} \right).
\end{aligned}
\]
% \vspace{-1mm}
As a result, the model's prediction for other classes is left unsupervised, which would otherwise serve as the third term in \cref{equ:three term loss}. 
This indicates a lack of supervision for the predictions of non-forgetting classes in the \textbf{retention loss}. 
Thus, some methods leverage remaining data to preserve performance on retained classes~\cite{salun}.

\vspace{-1mm}
\subsection{Mask Distillation Method}
\vspace{-1mm}
% \subsection{Decoupled Distillation Method}
\label{sec:our method}
% \notice{we further supervise non forget/ maybe ablation on previous work}
Previous work neglected the supervision of remaining classes, hindering the preservation of predictions for other classes. 
However, when forgetting specified classes, it is important to keep the predictions for other classes unchanged~\cite{boundary}. Additionally, preserving the relative magnitudes of the model’s outputs for other classes, which reflect its ``dark knowledge", also plays a crucial role~\cite{dkd}.
Thus, we aim to further retain the relative magnitudes of the non-forgotten outputs on \textbf{retention loss}, under the forgetting \condition. 
This motivates the use of distillation based on the original model's logits, rather than one-hot supervision, which tends to be overly rigid.

% \notice{with one assumption metioned in s4.3}
To summarize, our goal is to approximate the target distribution using three \textbf{key observations and \conditions}, building further on the \cref{sec:theoretical analysis}: 
\begin{itemize}
    \item \textbf{forgetting \condition}: The probability output for the unlearning class, \( p_u \), approaches zero
    % is an extremely small value
    , \ie, \( p_u = 0 \), as discussed in \cref{sec:theoretical analysis}.
    \item \textbf{retaining \condition}: The relative magnitude for the remaining classes remain unchanged, \ie, for \( i \neq u \), \( p_i \propto \mathrm{Softmax}(f_{\theta_{\mathrm o}}(\mathbf x))_i \).
    \item \textbf{probability property}: The target probability distribution must satisfy the basic property of a probability distribution, \ie, $\sum_{i=1}^kp_i=1$.
\end{itemize}

This inspires us to supervise the current unlearn model \( f_{\theta} \) leveraging the variant of distribution predicted by the frozen original model \( f_{\theta_{\mathrm o}} \). By applying a mask and vector normalization to the model output, we obtain an approximate target distribution that satisfies the above \conditions.

Let \( \mathbf{z}(\mathbf{x}; \theta_{\mathrm o}) = f_{\theta_{\mathrm o}}(\mathbf{x}) \) (simplified as \( \mathbf{z} \) hereafter) represent the logits of the frozen original model for $\mathbf{x}$,
where \( \mathrm{Softmax}(\mathbf{z}) \) represents the corresponding probability distribution. 
We introduce \( \mathrm{Mask}_u(\cdot) \) as a masking function applied to \( \mathbf{v} \in \mathbb{R}^K \), which sets the probability of class \( u \) to zero while leaving the other dimensions unchanged. Recalling that \( \mathbf{e}^{u} \) is the \( u \)-th column of the identity matrix \( I \), this masking function is defined as the Hadamard product:

\vspace{-3mm}
\[
\mathrm{Mask}_u(\mathbf{v}) = (\mathbf 1 -\mathbf{e}^{u}) \odot \mathbf{v}.
\]
% \vspace{-1mm}
The \( \mathrm{Normalize}(\cdot) \) function normalizes the vector to ensure that the sum of its elements equals 1, which is achieved by dividing each element by the total sum of all.

By applying the masking and normalization function to process the frozen model’s output as the target, we have:
\vspace{-2mm}
\begin{equation}
   \mathcal{L} = \mathrm{KL}\left( \mathrm{Normalize}(\mathrm{Mask}_u(\mathrm{Softmax}(\mathbf{z}))) \,\middle\|\, \mathbf q \right). 
   \label{equ:w/o norm loss}
\end{equation}

\vspace{-1mm}
The target in \cref{equ:w/o norm loss} satisfies all these \textbf{\conditions}. By masking out the probability for the target class, we satisfy \textbf{forgetting \condition}. The probabilities for the other classes remain after the masking process, and after normalization, they are proportional to the original model’s output probabilities, satisfying \textbf{retaining \condition}.
Clearly, the normalize function ensures the \textbf{probability property}.

Compared to explicitly specifying a one-hot target vector, the masking design allows for a more fine-grained utilization of the logits for the decoupled forgetting and non-forgetting components.
This approach not only enables effective unlearning but also preserves knowledge of the non-forgotten classes, as shown later in \cref{sec:ablation study}.
%, preventing excessive forgetting compared to the original model.

Moreover, we demonstrate that, with a carefully designed mask function $\mathrm{Mask}_u'(\cdot)$, the masking and softmax steps can be interchanged. This reordering allows the softmax vector to naturally sum to 1, reducing the need for normalization while achieving equivalent results. Further details are provided in the Appendix. %\notice{supp mat}
The resulting loss function, derived from \cref{equ:w/o norm loss}, is as follows:

\vspace{-3mm}
\[
\begin{aligned}
\mathcal{L} 
% &= \mathrm{KL}\left( \mathrm{Normalize}(\mathrm{Mask}_u(\mathrm{Softmax}(\mathbf z))) \,\middle\|\, \mathbf q \right) \\
% &= \mathrm{KL}\left( \mathrm{Normalize}(\mathrm{Softmax}(\mathrm{Mask}_u'(z))) \,\middle\|\,\mathbf q \right) \\
&= \mathrm{KL}\left( \mathrm{Softmax}(\mathrm{Mask}_u'(\mathbf z)) \,\middle\|\,\mathbf q \right).
\end{aligned}
\]
\vspace{-1mm}
For clarity, the Algorithm~\ref{alg:code} provides the final method.
% For clarity, the Algorithm \ref{alg:code} provides the pseudo-code for the final method.

\begin{algorithm}[t]
\caption{Pseudocode of DELETE in a PyTorch-like style.}
\label{alg:code}
% \algcomment{\fontsize{7.2pt}{0em}\selectfont \texttt{bmm}: batch matrix multiplication; \texttt{mm}: matrix multiplication; \texttt{cat}: concatenation.
% }
\definecolor{codeblue}{rgb}{0.25,0.5,0.5}
\lstset{
  backgroundcolor=\color{white},
  basicstyle=\fontsize{7.2pt}{7.2pt}\ttfamily\selectfont,
  columns=fullflexible,
  breaklines=true,
  captionpos=b,
  commentstyle=\fontsize{7.2pt}{7.2pt}\color{codeblue},
  keywordstyle=\fontsize{7.2pt}{7.2pt},
%  frame=tb,
}
\begin{lstlisting}[language=python]
# unlearn_model: the classification model to unlearn
# images: the input images to forget (N, 3, H, W)
# labels: the corresponding labels (N, K)

# Initialize the frozen model with unlearn model
frozen_model = copy(unlearn_model)

# Disable gradient computation for the frozen model
frozen_model.params.require_grad_(False)  

for images, labels in loader:
    # Create a decoupled distillation mask
    mask = zeros((N, K))
    mask[arange(N), labels] = -float("inf")
    
    # Compute soft targets using the frozen model
    frozen_outputs = frozen_model.forward(images)
    masked_outputs = mask + frozen_outputs
    soft_targets = softmax(masked_outputs, dim=1)  
    
    # Forward pass through the unlearn model
    logits = unlearn_model.forward(images)
    loss = KLDivLoss(softmax(logits, dim=1), soft_targets)
    
    # Update the unlearn model
    loss.backward()
    update(unlearn_model.params)
\end{lstlisting}
\vspace{-2mm}
\end{algorithm}

\vspace{-1mm}
\section{Experiments}
\vspace{-1mm}
\label{sec:exp}
% In this section, we first introduce the experiment setup and the evaluation metrics. 
% Then we systematically analyze the 3 properties of our methods: \sout{complete forgetting with perfect remaining ability, efficiency and illegal-oriented}. Finally, we perform extensive analysis to explore the potential of our models for wider applications.
\subsection{Experimental Setup}
\vspace{-1mm}

Considering privacy requirements, restricted access, computational constraints, and other real-world limitations as mentioned before, we impose two strict constraints in our experiments: \textbf{(\romannumeral 1)} No access to the remaining data; \textbf{(\romannumeral 2)} No intervention during the pre-training phase. 
Additional implementation details are provided in the appendix.

\noindent\textbf{Datasets.} 
To evaluate the performance on class-centric unlearning task, we use CIFAR-10, CIFAR-100~\cite{cifar}, and Tiny ImageNet~\cite{tinyimagenet} datasets. For both single-class and multi-class forgetting tasks, we randomly select a set of classes, then unlearn the same selected classes across methods.

\noindent\textbf{Models.} 
We adopt ResNet-18~\cite{resnet} as the primary architecture for forgetting experiments. Additionally, we evaluate methods on ViT-s~\cite{vit}, Swin-T~\cite{swin-t}, and VGGNet-16~\cite{vggnet} to assess effectiveness across different model structures.

\noindent\textbf{Baselines.} 
In our experiments, we compare the following baselines 
%that could unlearn without access to remaining data
: Random Label, Negative Gradient, Finetune, Fisher Forgetting~\cite{fisherforget}, Boundary Shrink~\cite{boundary}, Boundary Expand~\cite{boundary}, Learn to Unlearn~\cite{learn2unlearn}, Bad Teacher~\cite{badteacher}, Saliency Unlearn~\cite{salun}, SCRUB~\cite{scrub} and Influence Unlearn~\cite{influenceunlearn}. 

Additionally, we include the performance of the original model and the retrained model, with the latter often considered the gold standard in machine unlearning~\cite{boundary,necessity}.
% been published in the past year.

% We therefore focus on comparing the following methods. Random Label, Negative Gradient, Boundary Shrink, Boundary Expand, and Learn to Unlearn, which do not require remaining data. For Bad Teacher and Saliency Unlearn, we adapt them by removing remaining data related loss terms. We also include some parameter-scrubbing approaches, such as Fisher Forget and Influence Unlearning. Specially, we include the naive finetuning method, which leverages remaining data to achieve strong performance, highlighting the importance of further exploration into remain data free forgetting approaches.

% \noindent\textbf{Implementation details:}
% Our implementation is based on PyTorch, with all experiments conducted on one RTX 4090. 
% We pretrain the model for 150 epochs, then apply all unlearning methods on the pretrained model.
% \sout{Additional implementation details are provided in the appendix.}
% 为了贴近实际场景，我们使用了常见的数据增广策略，包括随机裁剪、水平翻转。其他的实现细节参考附录。

\noindent\textbf{Metrics.}
We employ the following accuracy to evaluate effectiveness:$\textbf{Acc}_{\textbf f}$ on the forget training data $\mathcal{D}_{\mathrm f}$, $\textbf{Acc}_{\textbf r}$ on the remain training data $\mathcal{D}_{\mathrm r}$, $\textbf{Acc}_{\textbf{ft}}$ on the forget test data $\mathcal{D}_{\mathrm{ft}}$, and $\textbf{Acc}_{\textbf{rt}}$ on the remain test data $\mathcal{D}_{\mathrm{rt}}$.

We further use $\textbf{H}\text{-}\textbf{Mean}$ as an overall performance measure of model forgetting and retention. Similar to gs-lora~\cite{gslora}, $\mathrm{H}$-$\mathrm{Mean}$ is computed as the harmonic mean of $\mathrm{Acc}_{\mathrm{rt}}$ and $\mathrm{Drop}_{\mathrm{ft}}$, where $\mathrm{Drop}_{\mathrm{ft}}$ represents the performance drop on the forget test data after unlearning. 

What's more, we employ \textbf{MIA}, the membership inference attack~\cite{mia} success rate, as another indicator of forgetting effectiveness. MIA measures the proportion of samples in $\mathcal D_{\mathrm f}$ that the unlearn model may still “memorize”, revealing potential remnants of the forget data~\cite{salun}. 
%\sout{See the appendix for further implementation details.}

% In class-centric unlearning tasks, the unlearn model’s performance should ideally align closely with that of the retrain model.
Metrics associated with \(\mathrm{Acc}_{\mathrm f}\), \(\mathrm{Acc}_{\mathrm{ft}}\), and MIA should approach zero to indicate effective forgetting, while metrics such as \(\mathrm{Acc}_{\mathrm r}\), \(\mathrm{Acc}_{\mathrm{rt}}\), and H-Mean are expected to align with original model's, demonstrating retention of knowledge in the remaining classes~\cite{boundary,salun,gslora}.

\vspace{-1mm}
\subsection{Results and Comparisons}
\vspace{-1mm}
% \notice{claim our method work well both toy and large scale dataset, some others fail}
% \subsection{Single Class Unlearning}
\noindent\textbf{Performance on Different Datasets.}
\begin{table*}[t]
    \centering
    \caption{Performance comparison of single-class forgetting across different unlearning methods on CIFAR-10 and CIFAR-100 datasets. 
    \g{Gray} indicates methods with remain data or intervention, and \textbf{bold} indicates the single best result among methods w/o remain data or intervention (if multiple, results are not in bold). The same notation applies hereafter.
    %The best results are highlighted in \textbf{bold}, while the second-best results are highlighted in \underline{underline}. Other tables follow the same notation.
    }
    \vspace{-3mm}
    \resizebox{1.7\columnwidth}{!}{
        \begin{tabular}{cc|cccccc|cccccc}
            \toprule
            \multicolumn{2}{c|}{\multirow{2}{*}{Method}} &
            \multicolumn{6}{c|}{CIFAR-10} & \multicolumn{6}{c}{CIFAR-100} \\
            \cmidrule(lr){3-8} \cmidrule(lr){9-14}
                &
                & $\mathrm{Acc}_{\mathrm{f}} \downarrow$ 
                & $\mathrm{Acc}_{\mathrm{r}} \uparrow$ 
                & $\mathrm{Acc}_{\mathrm{ft}} \downarrow$ 
                & $\mathrm{Acc}_{\mathrm{rt}} \uparrow$ 
                & $\mathrm{H\text{-}Mean} \uparrow$ 
                & $\mathrm{MIA} \downarrow$ 
                & $\mathrm{Acc}_{\mathrm{f}} \downarrow$ 
                & $\mathrm{Acc}_{\mathrm{r}} \uparrow$ 
                & $\mathrm{Acc}_{\mathrm{ft}} \downarrow$ 
                & $\mathrm{Acc}_{\mathrm{rt}} \uparrow$ 
                & $\mathrm{H\text{-}Mean} \uparrow$ 
                & $\mathrm{MIA} \downarrow$ \\
        \midrule
            \multicolumn{2}{c|}{Original Model} &100 & 100 & 97.00 & 95.31 & - & 99.08 & 100 & 99.98 & 67.00 & 77.79 & - & 98.00 \\
            \multicolumn{2}{c|}{Retrain Model} & 0 & 100 & 0 & 95.20 & 96.09 & 0 & 0 & 99.97 & 0 & 77.64 & 71.93 & 0\\
        \midrule
            \multicolumn{2}{c|}{Random Label} &\pz0.10 & 90.66 & \pz1.60 & 82.18 & 88.30 & 0 & \pz1.20 & 96.60 & \pz1.00 & 66.99 & 66.49 & 0\\
            \multicolumn{2}{c|}{Negative Gradient} & 11.12 & 83.94 & \pz7.20 & 75.52 & 82.04 & \pz7.44& \pz2.20 & 97.80 & 0 & 70.66 & 68.78 & \pz1.20\\
            \multicolumn{2}{c|}{Boundary Shrink} & \pz0.28 & 90.34 & \pz4.50 & 84.56 & 88.35 & \pz0.22 & \pz1.40 & 96.48 & 0 & 68.46 & 67.72 & 0\\
            \multicolumn{2}{c|}{Boundary Expand} &11.32 & 90.74 & 15.00 & 79.37 & 80.66 & \pz0.86 & \pz2.40 & 98.59 & \pz1.00 & 71.69 & 68.73 & 0\\
            \multicolumn{2}{c|}{Influence Unlearn} &\pz1.04 & 99.58 & \pz0.60 & 93.21 & 94.78 & 0 & 0 & 97.83 & 0 & 71.03 & 68.96 & 0\\
            \multicolumn{2}{c|}{Learn to Unlearn} &12.36 & 96.49 & \pz8.60 & 90.02 & 89.20 & \pz7.56 & \pz1.60 & 97.70 & 0 & 70.29 & 68.61 & \pz0.40\\
            \multicolumn{2}{c|}{Bad Teacher} &\pz1.98 & 92.23 & 18.80 & 87.13 & 82.42 & \pz0.52 & \pz1.60 & 98.46 & \pz3.00 & 71.38 & 67.49 & \pz0.20\\
            \multicolumn{2}{c|}{\g{Bad Teacher}} &\pz\g{0.34} & \g{99.99} & \pz\g{0.30} & \g{94.98} & \g{95.83} & \g{0} & \g{\pz0.40} & \g{99.98} & \g{0} & \g{77.63} & \g{71.92} & \g{0}\\
            \multicolumn{2}{c|}{Saliency Unlearn} &\pz9.98 & 79.19 & 10.50 & 70.18 & 80.53 & 11.70 & \pz1.20 & 88.07 & \pz3.00 & 59.73 & 61.79 & \pz1.00\\
            \multicolumn{2}{c|}{\g{Saliency Unlearn}} &\pz\g{0.84} & \g{99.73} & \pz\g{0.10} & \g{93.91} & \g{95.38} & \g{\pz0.02} & \g{\pz0.80} & \g{99.21} & \g{0} & \g{74.27} & \g{70.45} & \g{0}\\
            \multicolumn{2}{c|}{\g{Fisher Forget}} &\g{68.94} & \g{57.33} & \g{68.30} & \g{59.73} & \g{38.77} & \g{59.50} & \g{66.00} & \g{74.52} & \g{37.00} & \g{54.68} & \g{38.74} & \g{68.60}\\
            \multicolumn{2}{c|}{\g{SCRUB}} &\g{0} & \g{99.92} & \g{0} & \g{94.94} & \g{95.96} & \g{0} & \g{0} & \g{99.95} & \g{0} & \g{77.41} & \g{71.83} & \g{0}\\
            \multicolumn{2}{c|}{\g{Finetune}} &\pz\g{0.12} & \g{99.53} & \g{{0}} & \g{93.83} & \g{95.39} & \g{0} & \g{0} & \g{99.30} & \g{0} & \g{74.15} & \g{70.39} & \g{0}\\
        \midrule
            \multicolumn{2}{c|}{Ours} & \textbf{0} & \textbf{100} & \textbf{0} & \textbf{95.03} & \textbf{96.00} & 0 & 0 & \textbf{99.97} & 0 & \textbf{76.57} & \textbf{71.47} & 0\\
            % \multicolumn{2}{c|}{Learn to Unlearn(IMP)} & - & - & - & - & - & - & - & - & - & - & - & -\\
            % \multicolumn{2}{c|}{Unrolling} & - & - & - & - & - & - & - & - & - & - & - & -\\
            % 需要remain数据
            % \multicolumn{2}{c|}{SCRUB} & - & - & - & - & - & - & - & - & - & - & - & -\\
            % \multicolumn{2}{c|}{SSD} & - & - & - & - & - & - & - & - & - & - & - & -\\ % 需要remain数据
            % 需要特殊结构
            % \multicolumn{2}{c|}{amnesiac} & - & - & - & - & - & - & - & - & - & - & - & -\\
            % \multicolumn{2}{c|}{SISA} & - & - & - & - & - & - & - & - & - & - & - & -\\
            % 需要repair阶段
            % \multicolumn{2}{c|}{UNSIR} & - & - & - & - & - & - & - & - & - & - & - & -\\ 
            % \multicolumn{2}{c|}{LAF} & - & - & - & - & - & - & - & - & - & - & - & -\\
            
            % \multicolumn{2}{c|}{xxx} & - & - & - & - & - & - & - & - & - & - & - & -\\
            \bottomrule
        \end{tabular}
    }
    \label{tab:cifar10 cifar100}
    \vspace{-4mm}
\end{table*}
In this part, we discuss the single-class forgetting performance across various datasets. 
Although our experimental setup does not allow access to remaining data or intervention, we still compare several methods that have these, marking them with \textcolor{gray}{gray}.
% An effective forgetting algorithm should remove target class knowledge while maintaining classification accuracy on other classes, comparable to that of the retrain model. 

Using CIFAR-10 as an example in \cref{tab:cifar10 cifar100}, although all methods achieve varying degrees of forgetting in $\mathrm{Acc}_{\mathrm{ft}}$, 
some methods like Negative Gradient maintain $\mathrm{Acc}_{\mathrm{f}}$ and $\mathrm{Acc}_{\mathrm{ft}}$ around 10\%, indicating incomplete forgetting. 
% methods such as Negative Gradient, Boundary Expand, and Learn to Unlearn maintain $\mathrm{Acc}_{\mathrm{f}}$ and $\mathrm{Acc}_{\mathrm{ft}}$ around 10\%, indicating incomplete forgetting. 
In contrast, our method achieves complete forgetting in $\mathrm{Acc}_{\mathrm{ft}}$ while maintaining an excellent $\mathrm{Acc}_{\mathrm{rt}}$, with a minimal gap of only 0.17\% compared to the retrain model. 
When experiments are extended to larger datasets such as CIFAR-100 and Tiny ImageNet (\cref{tab:cifar10 cifar100} and Appendix),
%\cref{tab:tiny imagenet} 
the performance of some methods fluctuates significantly. 
%For instance, Learn to Unlearn drops from third place on CIFAR-10 to the lowest rank on Tiny ImageNet, revealing its fragility across datasets. 
Notably, our method consistently outperforms all other methods across all datasets and metrics.

\noindent\textbf{Performance on Different Models.}
\begin{table}[t]
    \centering
    \caption{Performance comparison across different models.}
    \vspace{-3mm}
    \resizebox{\columnwidth}{!}{
        \begin{tabular}{cc|cccccc}
            \toprule
            % \cmidrule(lr){3-8}
                \multicolumn{2}{c}{}  
                & $\mathrm{Acc}_{\mathrm{f}} \downarrow$ 
                & $\mathrm{Acc}_{\mathrm{r}} \uparrow$ 
                & $\mathrm{Acc}_{\mathrm{ft}} \downarrow$ 
                & $\mathrm{Acc}_{\mathrm{rt}} \uparrow$ 
                & $\mathrm{H\text{-}Mean} \uparrow$ 
                & $\mathrm{MIA} \downarrow$ \\
            \midrule
            \multicolumn{8}{l}{\emph{VGGNet-16}\vspace{0.02in}}\\
            \multicolumn{2}{l}{\pz\pz Original Model} & 99.94 & 99.94 & 91.20 & 92.02 & - & 99.88 \\
            \multicolumn{2}{l}{\pz\pz Retrain Model} & 0 & 99.71 & 0 & 92.49 & 91.84 & 0 \\
            \multicolumn{2}{l}{\pz\pz Boundary Shrink} & \pz6.68 & 91.15 & \pz6.80 & 82.79 & 83.59 &\pz2.84\\
            \multicolumn{2}{l}{\pz\pz Ours} & 0 & 99.66 & 0 & 92.08 & 91.64 & 0 \\
            \midrule
            \multicolumn{8}{l}{\emph{Swin-T}\vspace{0.02in}}\\
            \multicolumn{2}{l}{\pz\pz Original Model} & 85.00 & 86.58 & 82.70 & 82.68 & - & 82.94 \\
            \multicolumn{2}{l}{\pz\pz Retrain Model} & 0 & 86.41 & 0 & 82.77 & 82.73 & 0 \\
            \multicolumn{2}{l}{\pz\pz Boundary Shrink} & \pz5.54 & 62.94 & \pz2.50 & 58.18 & 67.44 & 11.36 \\
            \multicolumn{2}{l}{\pz\pz Ours} & 0 & 85.37 & 0 & 83.09 & 82.89 & 0 \\
            \midrule
            \multicolumn{8}{l}{\emph{ViT-S}\vspace{0.02in}}\\
            \multicolumn{2}{l}{\pz\pz Original Model} & 90.50 & 90.57 & 72.70 & 75.73 & - & 84.60 \\
            \multicolumn{2}{l}{\pz\pz Retrain Model} & 0 & 92.42 & 0 & 76.44 & 74.52 & 0 \\
            \multicolumn{2}{l}{\pz\pz Boundary Shrink} & \pz1.84 & 73.16 & \pz1.70 & 65.59 & 68.19 & \pz1.66 \\
            \multicolumn{2}{l}{\pz\pz Ours} & 0 & 90.74 & 0 & 77.21 & 74.89 & 0 \\
            \bottomrule
        \end{tabular}
    }
    \vspace{-4mm}
    \label{tab:model}
\end{table}
% To evaluate effectiveness across various models, we conduct single-class unlearning experiments on CIFAR-10 using three other models: VGGNet-16, Swin-T, and Vit-S, as shown in \cref{tab:model}. A full comparison of methods is provided in the Appendix.
To evaluate effectiveness across various models, we further conduct single-class unlearning experiments with VGGNet-16, Swin-T, and Vit-S, as shown in \cref{tab:model}. A full comparison of methods is provided in the Appendix.
At the same performance of $\mathrm{Acc}_{\mathrm{ft}}$, our method shows a maximum performance gap of only 0.77\% in $\mathrm{Acc}_{\mathrm{rt}}$ compared to the retrain model, demonstrating consistently excellent results across different architectures while Boundary Shrink shows limited performance, validating the robustness of DELETE.

% with remaining accuracy on Swin-T dropping from 82.68\% to 58.18\%.

% In contrast, the performance of Boundary Shrink fluctuates. 
% Specifically, on the Swin-T model, Boundary Shrink’s remaining accuracy drops significantly from 82.77\% to 58.18\%, indicating limited adaptability to Swin-T.
% While Boundary Shrink generally maintains a good $\mathrm{Acc}_{\mathrm{ft}}$, it exhibits significant variability in $\mathrm{Acc}_{\mathrm{rt}}$. 

\noindent\textbf{Performance with Frozen Linear Classifier.}
% \subsection{Unlearning on Feature Exactor under Freeze Linear Classifier Setup}
\begin{table}[t]
    \centering
    \caption{Performance comparison with frozen linear classifier.}
    \vspace{-3mm}
    \resizebox{\columnwidth}{!}{
        \begin{tabular}{cc|cccccc}
            \toprule
            % \cmidrule(lr){3-8}
                \multicolumn{2}{c|}{\multirow{1}{*}{Method}}  
                & $\mathrm{Acc}_{\mathrm{f}} \downarrow$ 
                & $\mathrm{Acc}_{\mathrm{r}} \uparrow$ 
                & $\mathrm{Acc}_{\mathrm{ft}} \downarrow$ 
                & $\mathrm{Acc}_{\mathrm{rt}} \uparrow$ 
                & $\mathrm{H\text{-}Mean} \uparrow$ 
                & $\mathrm{MIA} \downarrow$ \\
            \midrule
            \multicolumn{2}{c|}{Original Model} & 100 & 100 & 97.00 & 95.31 & - & 99.08 \\
            \multicolumn{2}{c|}{Retrain Model} & 0 & 100 & 0 & 95.12 & 96.05 & 0 \\
            \midrule
            \multicolumn{2}{c|}{Random Label} & \pz0.82 & 93.60 & \pz6.90 & 85.51 & 87.75 & 0 \\
            % \multicolumn{2}{c|}{Finetune} & \pz2.72 & 99.84 & \pz2.40 & 94.19 & 94.39 & \pz0.02 \\
            \multicolumn{2}{c|}{Negative Gradient} & 19.20 & 87.83 & 15.00 & 79.37 & 80.66 & 13.88 \\
            \multicolumn{2}{c|}{Boundary Shrink} & \pz1.12 & 87.97 & \pz9.30 & 82.43 & 84.98 & \pz1.36 \\
            \multicolumn{2}{c|}{Boundary Expand} & 58.44 & 32.02 & 69.60 & 39.13 & 32.23 & 66.62 \\
            \multicolumn{2}{c|}{Bad Teacher} & \pz4.14 & 86.26 & 24.80 & 81.51 & 76.57 & \pz2.14 \\
            \multicolumn{2}{c|}{Saliency Unlearn} & 10.58 & 85.95 & 11.20 & 76.14 & 80.68 & \pz3.78 \\
            % \multicolumn{2}{c|}{Fisher Forget} & 55.32 & 55.36 & 54.10 & 55.37 & 48.34 & 51.98 \\
            \multicolumn{2}{c|}{Influence Unlearn} & \pz0.40 & 99.61 & \pz0.20 & 93.59 & 95.17 & \pz0.20 \\
            \multicolumn{2}{c|}{Learn to Unlearn} & 21.00 & 97.86 & 17.60 & 91.09 & 84.84 & 15.54 \\
            \midrule
            \multicolumn{2}{c|}{Ours} & \textbf{0} & \textbf{100} & \textbf{0} & \textbf{95.44} & \textbf{96.21} & 0 \\
            \bottomrule
        \end{tabular}
    }
    \label{tab:freeze linear}
    \vspace{-4mm}
\end{table}
A key concern in machine unlearning is whether knowledge of the target class is truly forgotten. Studies suggest that forgetting often occurs in the model’s linear classifier~\cite{learn2unlearn}. However, merely modifying classifier to misclassify without erasing knowledge at the feature level doesn't fulfill the goal of machine unlearning~\cite{fisherforget}. Incomplete forgetting within the feature extractor could also introduce risks of privacy leakage.

To address this, we freeze the model's linear classifier and train only the feature extractor to forget. Results, shown in \cref{tab:freeze linear}, demonstrate that our method consistently outperforms existing approaches across all metrics.
Boundary Expand shows a significant drop in overall performance, suggesting such methods \textbf{achieve unlearning primarily at the classifier level rather than at the feature level}.
% However, Boundary Expand shows a substantial drop in overall performance, with Acc\(_{\text{ft}}\) degrading from 15.00\% to 69.60\% and Acc\(_{\text{rt}}\) from 79.37\% to 39.13\%.
% This results suggests that such manipulations achieve unlearning primarily by enforcing misclassification at the classifier level rather than inducing feature-level forgetting.

% In this frozen-classifier setup, the performance of certain methods degrades considerably.
% For example, Boundary Expand shows a substantial drop in overall performance, with Acc\(_{\text{ft}}\) degrading from 15.00\% to 69.60\% and Acc\(_{\text{rt}}\) from 79.37\% to 39.13\%, due to its reliance on manipulations specific to the linear classifier, such as expanding and pruning ``shadow neurons". 

% This results suggests that such manipulations achieve unlearning primarily by enforcing misclassification at the classifier level rather than inducing feature-level forgetting.
% Additional, for Negative Gradient, both $\mathrm{Acc}_{\mathrm{ft}}$ and MIA double to 15.00\% and 13.88\%, respectively, indicating incomplete forgetting at the feature extractor.

% \subsection{Multiple Classes Unlearning}
\noindent\textbf{Performance on Multi-Class Forgetting.}
\begin{table*}[t]
    \centering
    \caption{Comparison of the classification accuracy of multi-class forgetting across methods on CIFAR-100 dataset.}
    \vspace{-3mm}
    \resizebox{2\columnwidth}{!}{
       \begin{tabular}{cc|ccc|ccc|ccc|ccc|ccc}
            \toprule
            \multicolumn{2}{c|}{\multirow{2}{*}{Method}} & 
            \multicolumn{3}{c|}{1 Class} & 
            \multicolumn{3}{c|}{2 Classes} & 
            \multicolumn{3}{c|}{5 Classes} & 
            \multicolumn{3}{c|}{10 Classes} & 
            \multicolumn{3}{c}{20 Classes} \\
            \cmidrule(lr){3-5} \cmidrule(lr){6-8} \cmidrule(lr){9-11} \cmidrule(lr){12-14} \cmidrule(lr){15-17}
             & 
                & $\mathrm{Acc}_{\mathrm{ft}} \downarrow$ 
                & $\mathrm{Acc}_{\mathrm{rt}} \uparrow$ 
                & $\mathrm{H\text{-}Mean} \uparrow$ 
                % & \multicolumn{3}{c|}{}  
                & $\mathrm{Acc}_{\mathrm{ft}} \downarrow$ 
                & $\mathrm{Acc}_{\mathrm{rt}} \uparrow$ 
                & $\mathrm{H\text{-}Mean} \uparrow$ 
                % & \multicolumn{3}{c|}{} 
                & $\mathrm{Acc}_{\mathrm{ft}} \downarrow$ 
                & $\mathrm{Acc}_{\mathrm{rt}} \uparrow$ 
                & $\mathrm{H\text{-}Mean} \uparrow$ 
                % & \multicolumn{3}{c|}{} 
                & $\mathrm{Acc}_{\mathrm{ft}} \downarrow$ 
                & $\mathrm{Acc}_{\mathrm{rt}} \uparrow$ 
                & $\mathrm{H\text{-}Mean} \uparrow$ 
                % & \multicolumn{3}{c|}{} 
                & $\mathrm{Acc}_{\mathrm{ft}} \downarrow$ 
                & $\mathrm{Acc}_{\mathrm{rt}} \uparrow$ 
                & $\mathrm{H\text{-}Mean} \uparrow$ \\
            \midrule 
            \multicolumn{2}{c|}{Original Model}     & 67.00 & 77.79 & -     & 78.00 & 77.66 & -     & 76.80 & 77.72 & -     & 78.80 & 77.54 & -     & 77.40 & 77.74 & -     \\
            \multicolumn{2}{c|}{Retrain Model}      & 0     & 77.64 & 71.93 & 0     & 78.09 & 78.04 & 0     & 78.63 & 77.70 & 0     & 79.08 & 78.94 & 0     & 80.35 & 78.85 \\
            \midrule
            \multicolumn{2}{c|}{Random Label}       & 1.00  & 66.99 & 66.49 & 4.00  & 65.97 & 69.75 & 2.80  & 63.51 & 68.35 & 3.10  & 59.97 & 66.92 & 4.45  & 59.35 & 65.45 \\
            \multicolumn{2}{c|}{Negative Gradient}  & 0     & 70.66 & 68.78 & 1.50  & 69.96 & 73.08 & 0.20  & 68.97 & 72.59 & 2.10  & 67.11 & 71.59 & 3.10  & 40.60 & 52.51 \\
            \multicolumn{2}{c|}{Boundary Shrink}    & 0     & 68.46 & 67.72 & 1.00  & 60.10 & 67.51 & 0.40  & 63.68 & 69.46 & 0.50  & 59.19 & 67.42 & 3.90  & 61.90 & 67.20 \\
            \multicolumn{2}{c|}{Boundary Expand}    & 1.00  & 71.69 & 68.73 & 0.50  & 69.00 & 73.00 & 2.80  & 69.03 & 71.43 & 2.70  & 64.79 & 69.99 & 5.80  & 62.59 & 66.79 \\
            \multicolumn{2}{c|}{Influence Unlearn}  & 0     & 71.03 & 68.96 & 0     & 67.84 & 72.57 & 2.20  & 62.33 & 67.92 &\textbf{0}& 37.56&50.87& 20.20 & 43.77 & 49.59 \\
            \multicolumn{2}{c|}{Learn to Unlearn}   & 0     & 70.29 & 68.61 & 0.50  & 69.35 & 73.20 & 0     & 58.80 & 66.61 & 1.80  & 66.56 & 71.40 & 20.65 & 64.90 & 60.55 \\
            \multicolumn{2}{c|}{Bad Teacher}        & 3.00  & 71.38 & 67.49 & 0     & 59.48 & 67.49 & 0.80  & 62.61 & 68.66 & 0.50  & 58.17 & 66.75 & 3.10  & 55.07 & 63.26 \\
            \multicolumn{2}{c|}{Saliency Unlearn}   & 3.00  & 59.73 & 61.79 & 2.50  & 56.60 & 64.70 & 3.00  & 64.64 & 68.92 & 3.20  & 53.84 & 62.89 & 6.05  & 62.31 & 66.52 \\
            \midrule
            \multicolumn{2}{c|}{Ours} & 0 & \textbf{76.57} & \textbf{71.47} & 0 & \textbf{76.78} & \textbf{77.39} & 0 & \textbf{77.76} & \textbf{77.28} & 0.50 & \textbf{77.61} & \textbf{77.95} & \textbf{1.20} & \textbf{79.30} & \textbf{77.72}  \\
            % \multicolumn{2}{c|}{Fisher Forget} & 37.00 & 54.68 & 38.74  & - & - & - & - & - & - & - & - & - & - & - & -  \\
            % \multicolumn{2}{c|}{Finetune} & 0 & 74.15 & 70.39 & 0 & 75.22 & 70.87 & 0 & 75.36 & 70.93 & 0.20 & 75.50 & 70.88 & 0.05 & 76.94 & 71.60  \\
            % \multicolumn{2}{c|}{Learn to Unlearn(IMP)} & - & - & - & - & - & - & - & - & - & - & - & - & - & - & -  \\
            % \multicolumn{2}{c|}{SCRUB} & - & - & - & - & - & - & - & - & - & - & - & - & - & - & -  \\
            \bottomrule
        \end{tabular}    
    }
    \label{tab:multi class}
    \vspace{-4mm}
\end{table*}
Building on single-class forgetting in CIFAR-100, we further experiment with forgetting multiple classes. 
As shown in \cref{tab:multi class}, the results indicate that some methods suffer significant performance drops as forgetting class number increases. 
%For example, in terms of overall $\textrm{H}\text{-}\textrm{Mean}$ performance, Influence Unlearn, which is the second-best method after ours in single-class forgetting, becomes the worst-performing method when the classes number increases to 20. Its performance drops from 68.96\% to 45.20\%, demonstrating extreme sensitivity to the number of forgotten classes. Similarly, Negative Gradient also experiences a catastrophic drop, from 68.78\% to 49.65\% in H\text{-}Mean.
%Furthermore, all other methods exhibit notable performance degradation with multiple classes. 
When forgetting 20 classes, none of the other methods achieve both an $\mathrm{Acc}_{\mathrm{ft}}$ of 0\% and an $\mathrm{Acc}_{\mathrm{rt}}$ above 65\%, as they do in single-class forgetting. 
In contrast, our method demonstrates exceptional multi-class forgetting capability, showing the power of our novel loss function.

% with a minimal performance gap of only 1.15\% in $\textrm{H}\text{-}\textrm{Mean}$ compared to the retrain model when forgetting 20 classes, whereas other methods exhibit gaps of at least 10.58\%.

\begin{figure}[t]
    \centering
    \includegraphics[width=\columnwidth]{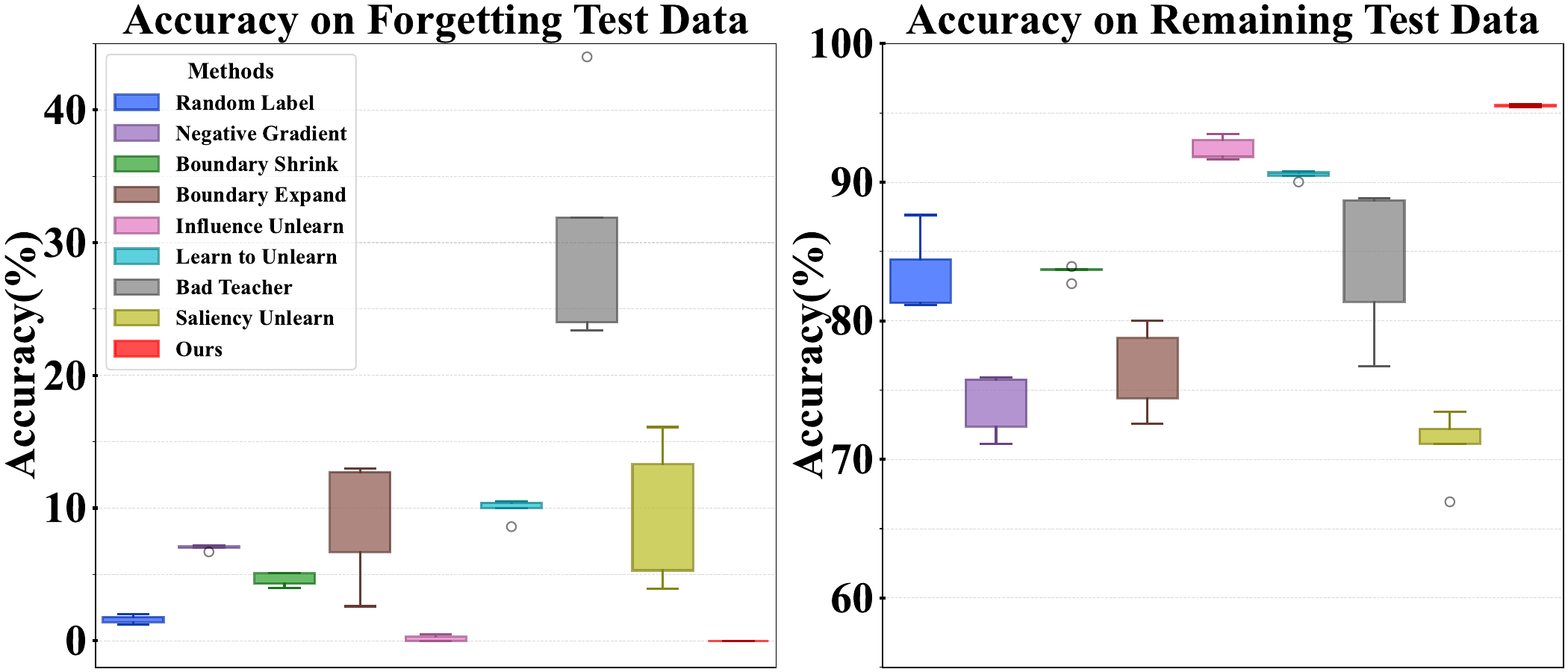}
    \vspace{-6mm}
    \caption{
    Boxplot comparison of accuracy on $\mathcal{D}_{\textrm{ft}}$ (\textbf{left}) and $\mathcal{D}_{\textrm{rt}}$ (\textbf{right}) across 5 runs, illustrating accuracy distributions and stability. \textcolor{red}{Ours} consistently achieves the lowest $\textrm{Acc}_{\textrm{ft}}$ and the highest $\textrm{Acc}_{\textrm{rt}}$, demonstrating stable performance across executions.
    }
    \label{fig:multi times runs}
    % \vspace{-1mm}
\end{figure}

\noindent\textbf{Performance Comparison of Stability and Insensitivity.}
In this part, we discuss the stability and learning rate insensitivity of different methods.
To evaluate the stability of each method, we conduct five repeated single-class forgetting runs. 
As shown in \cref{fig:multi times runs}, some methods exhibit considerable performance variance, indicated by the width of the boxplots. 
%For example, Bad Teacher method shows substantial variance, due to inconsistent performance of each run’s random incompetent teacher. In contrast, other methods, such as Learn to Unlearn, Influence Unlearn, and 
Meanwhile, ours demonstrate excellent stability.
%, with a clear advantage in Acc.

\begin{figure}[t]
    \centering
    \vspace{-3mm}
    \includegraphics[width=\columnwidth]{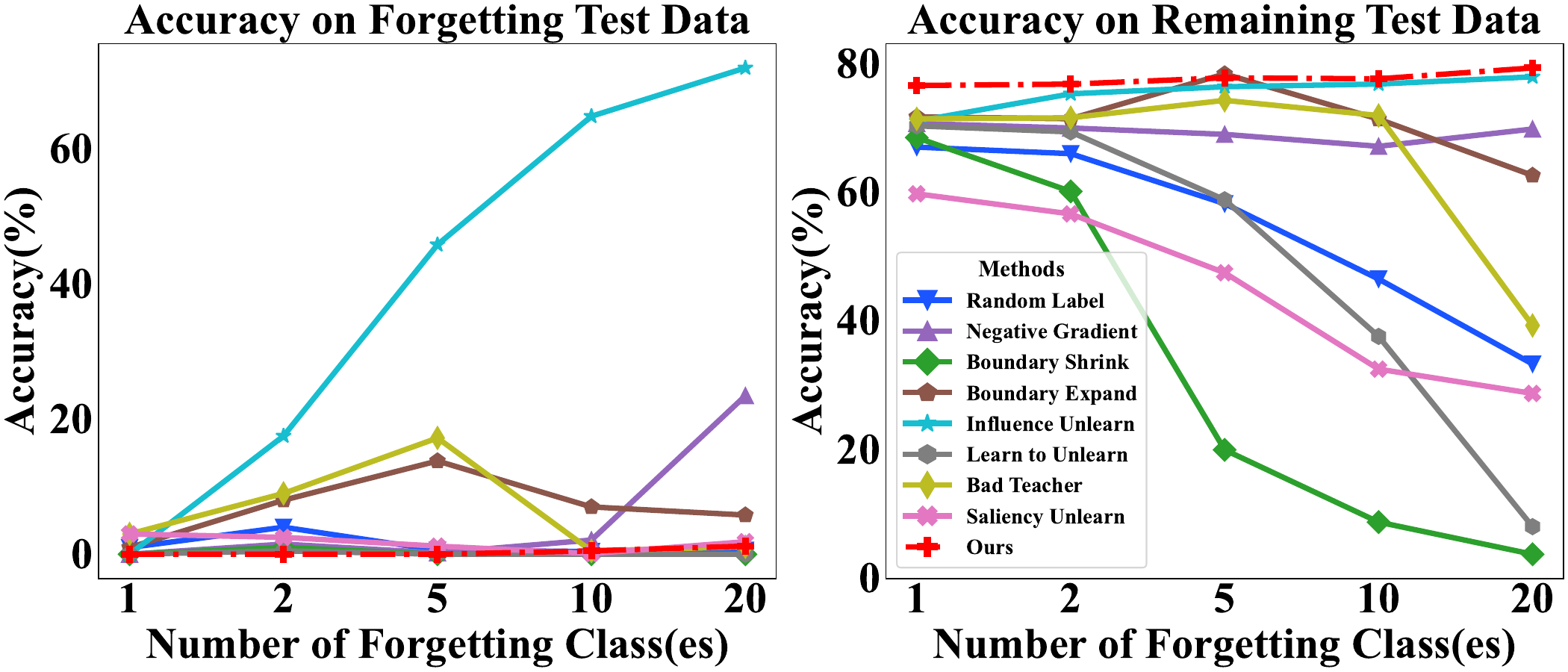}
    \vspace{-7mm}
    \caption{Accuracy comparison on $\mathcal{D}_{\textrm{ft}}$ (\textbf{left}) and $\mathcal{D}_{\textrm{rt}}$ (\textbf{right}) across multiple forgetting classes with a fixed learning rate. \textcolor{red}{Ours} achieves superior $\textrm{Acc}_{\textrm{ft}}$ and $\textrm{Acc}_{\textrm{rt}}$, whereas all other methods exhibit varying degrees of performance degradation.}
    \label{fig:multi class with fixed lr}
    \vspace{-4mm}
\end{figure}

Moreover, an promising method is expected to demonstrate insensitivity to the learning rate across varying numbers of forgetting classes, \ie, achieving strong performance with a fixed learning rate. 
% Otherwise, forgetting different classes would require frequent adjustments, resulting in additional computational overhead and unstable performance.
Otherwise, a different number of unlearning classes would require frequent adjustments, resulting in additional computational overhead and unstable performance.
As shown in \cref{fig:multi class with fixed lr}, our method achieves effective forgetting while preserving high accuracy on the remaining data. In contrast, all other methods exhibit varying degrees of performance degradation in $\textrm{Acc}_{\textrm{rt}}$ or $\textrm{Acc}_{\textrm{ft}}$, resulting in a noticeable performance gap compared to ours.

% As shown in \cref{fig:multi class with fixed lr}, our method achieves effective forgetting while preserving high accuracy on the remaining data. In contrast, all other methods exhibit varying degrees of performance degradation in $\textrm{Acc}_{\textrm{rt}}$, resulting in a noticeable performance gap compared to our approach, except for the Influence Unlearn. However, Influence Unlearning fails to achieve effective unlearn on $\textrm{Acc}_{\textrm{ft}}$.

% Other methods display greater fluctuation, highlighting the robustness of our approach in achieving effective unlearning while retaining essential knowledge.

\vspace{-1mm}
\subsection{Ablation Study}
% \vspace{-1mm}
\label{sec:ablation study}
In this part, we conduct ablation studies to validate the effectiveness of the forgetting and retention \conditions.

For the forgetting \condition, the output on forget class is required to approach 0, enforcing the erasure of target class. To validate this, we define \( p_u = \alpha \cdot \mathrm{Softmax}(f_{\theta_{\mathrm{o}}}(\mathbf{x}))_u \), where setting \( \alpha = 0 \) represents our default setting. 
Increasing \( \alpha \) deviates from this, and is expected to gradually weaken the forgetting effect.

Similarly, the retention \condition\ is designed to preserve non-target classes knowledge. It's implemented with \( p_i \propto \mathrm{Softmax}(f_{\theta_{\mathrm{o}}}(\mathbf{x})/T)_i \), where \( T = 1 \) is our default setting.
Increasing \( T \) deviates from the retention \condition\ and is anticipated to reduce the retention effect.
%in the accuracy of other categories.

% For the forgetting condition, our approach requires the model output on forget class to approach zero, effectively enforcing the erasure of target class. To validate this, we define \( p_u = \alpha \cdot \mathrm{Softmax}(f_{\theta_{\mathrm{o}}}(\mathbf{x}))_u \), where setting \( \alpha = 0 \) represents our default setting. 
% Larger values of \( \alpha \) would weaken the forgetting condition, which is expected to gradually diminish the forgetting effect.

% Similarly, the retention condition is designed to ensure that outputs for other categories remain stable, preserving non-target information. This is implemented with \( p_i \propto \mathrm{Softmax}(f_{\theta_{\mathrm{o}}}(\mathbf{x})/T)_i \), where \( T = 1 \) is our default setting.
% Increasing \( T \) deviates from the retention condition and is anticipated to cause a substantial decrease in the accuracy of other categories.

\begin{table}[t]
    \centering
    \caption{Ablation study on forgetting and retention \conditions, with varying values of $\alpha$ and $T$ respectively.}
    \vspace{-3mm}
    \resizebox{0.8\columnwidth}{!}{
        \begin{tabular}{ccccccc}
            \toprule
                \multicolumn{2}{c}{%Method
                }  
                & $\mathrm{Acc}_{\mathrm{f}} \downarrow$ 
                & $\mathrm{Acc}_{\mathrm{r}} \uparrow$ 
                & $\mathrm{Acc}_{\mathrm{ft}} \downarrow$ 
                & $\mathrm{Acc}_{\mathrm{rt}} \uparrow$ 
                & $\mathrm{H\text{-}Mean} \uparrow$ \\
            \midrule
            \multicolumn{2}{c}{Retrain Model} & 0 & 100 & 0 & 95.20 & 96.09  \\
            \multicolumn{2}{c}{Our Method} & 0 & 100 & 0 & 95.03 & 96.00  \\
            \midrule
            \multicolumn{7}{l}{\emph{Effect of varying $\alpha$}\vspace{0.02in}} \\
            % \midrule
            \multicolumn{2}{l}{\pz\pz $\alpha = 0.25$} & 41.00 & 100   & 37.30 & 95.49 & 73.47 \\
            \multicolumn{2}{l}{\pz\pz $\alpha = 0.50$} & 91.84 & 99.99 & 76.80 & 95.31 & 33.33 \\
            \multicolumn{2}{l}{\pz\pz $\alpha = 0.75$} & 99.88 & 99.97 & 95.00 & 94.36 & \pz3.92  \\
            \midrule
            \multicolumn{7}{l}{\emph{Effect of varying $T$} \vspace{0.02in}} \\
            % \midrule
            \multicolumn{2}{l}{\pz\pz $T = 5$} & 0 & 78.60 & 0 & 76.92 & 85.80 \\
            \multicolumn{2}{l}{\pz\pz $T = 10$} & 0 & 30.13 & 0 & 30.38 & 46.27 \\
            \multicolumn{2}{l}{\pz\pz $T = 15$} & 0 & 17.51 & 0 & 18.81 & 31.51 \\
            \bottomrule
        \end{tabular}
    }
    \label{tab:ablation study}
    \vspace{-2mm}
\end{table}

The experimental results, as shown in \cref{tab:ablation study}, align well with our expectations. When progressively increasing \( \alpha \) to violate the forgetting \condition, the model's accuracy on \( \mathcal{D}_{\mathrm{ft}} \) steadily rises from  0\%  to  95.00\%.
Likewise, increasing \( T \) to deviate from the retention \condition\ reduces the model's accuracy on \( \mathcal{D}_{\mathrm{rt}} \), from 95.03\%  to 18.81\%.
It is noteworthy that when varying \( \alpha \) to validate the forgetting \condition, $\text{Acc}_\text{rt}$ remains stable, while varying \( T \) results in the same stability in $\text{Acc}_\text{ft}$. 
This confirms the validity of our \conditions\ in effectively controlling forgetting without compromising retention, and vice versa.

\vspace{-1mm}
\subsection{Feature Representation Visualization}
% \vspace{-1mm}
% \noindent\textbf{Feature Representation Visualization:}
\begin{figure}[t]
    \centering
    \includegraphics[width=\columnwidth]{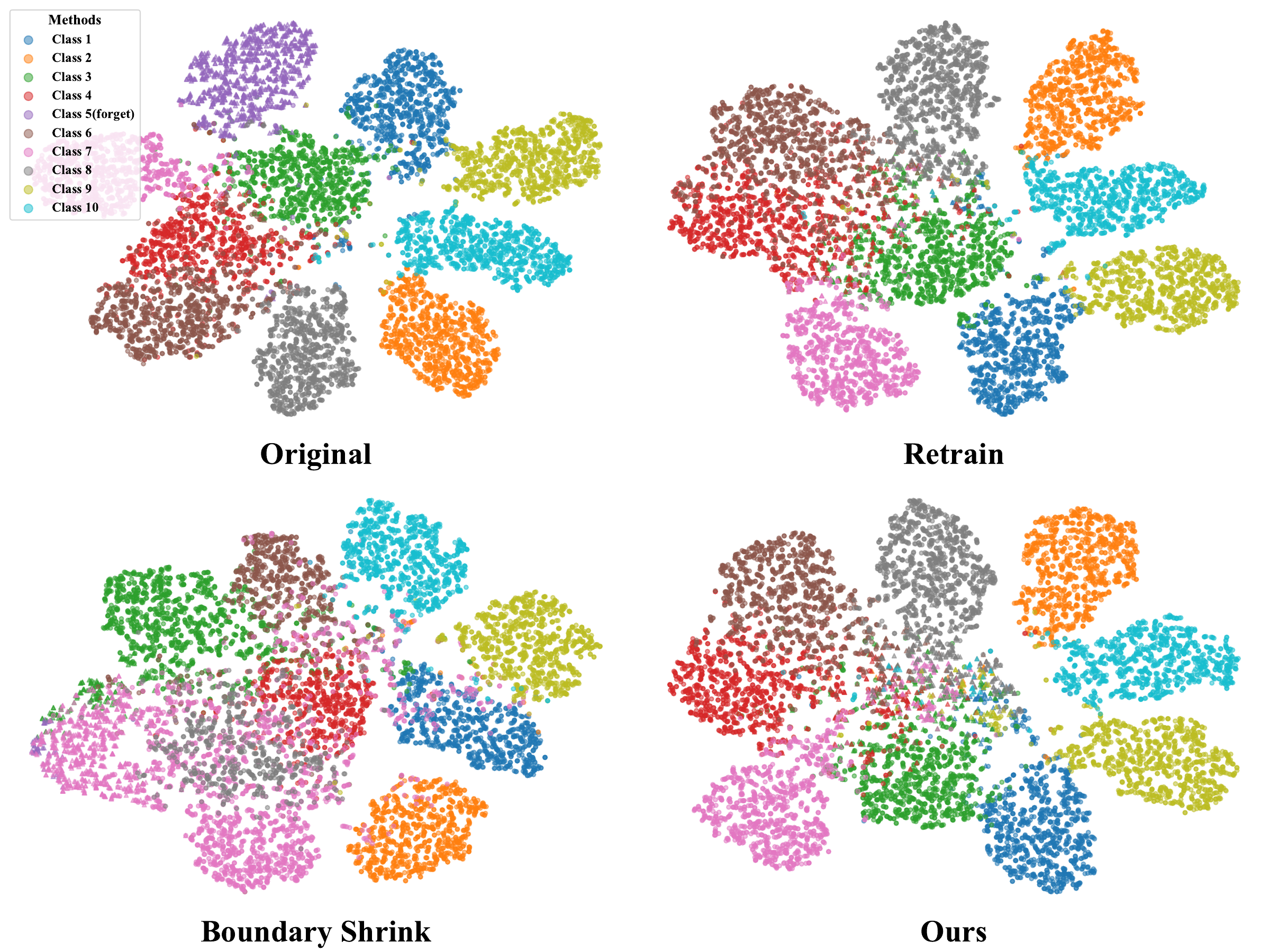}
    \vspace{-7mm}
    \caption{T-SNE visualization of feature representations space. 
    %Different colors indicate the predicted labels of samples. Specifically, triangles represent the forgetting class samples. 
    %Both ours and the retrain model successfully unlearn the triangle-labeled class while preserving the boundaries of other classes.
    }
    \label{fig:tsne}
    \vspace{-4mm}
\end{figure}

To investigate the impact of forgetting on model's feature space, we use t-SNE~\cite{tsne} to visualize the feature distributions. Different colors indicate distinct predicted classes, and triangles are used to mark samples belonging to the forgetting class. 
%An ideal method should effectively forget target classes while avoiding negative interference with other classes.
% As illustrated in \cref{fig:tsne}, the original model exhibits clear class boundaries. Both ours and retrain model successfully unlearn the \textcolor{custompurple}{purple} triangle-labeled samples, while preserving the boundaries of the remaining classes.
As illustrated in \cref{fig:tsne}, the original model exhibits clear class boundaries. Both ours and retrain model successfully unlearn the forget class samples, marked in \textcolor{custompurple}{purple}, while preserving the boundaries of the remaining classes.
In contrast, Boundary Shrink causes noticeable boundary degradation in certain classes; for example, the \textcolor{customPink}{pink} dots are dispersed across various regions of the space.

\vspace{-1mm}
\section{Application to Downstream Tasks}
\vspace{-1mm}
% In this section, we explore applying machine unlearning to downstream tasks such as face recognition, backdoor defense, and semantic segmentation, focusing on its potential for privacy protection and model correction. Additional implementation details are in the Appendix.
% In this section, we explore applying machine unlearning to downstream tasks, focusing on its potential for privacy protection and model correction. The implementation details are provided in the Appendix.
% This section explores the potential of machine unlearning for downstream tasks. See Appendix for more details.
% \vspace{-2mm}
\subsection{Face Recognition with Unlearning}
% \vspace{-1mm}
\begin{figure}[t]
    \centering
    \includegraphics[width=\columnwidth]{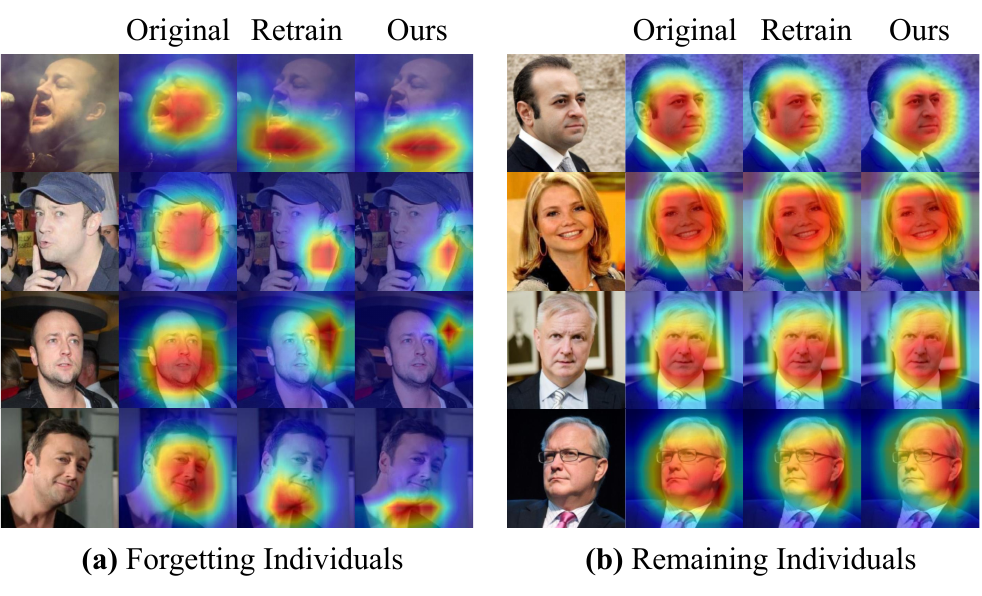}
    \vspace{-8mm}
    \caption{Grad-CAM visualization of face recognition models on forgetting and remaining individuals.}
    \label{fig:gradcam}
    \vspace{-4mm}
\end{figure}

\begin{figure}[t]
    \centering
    \includegraphics[width=\columnwidth]{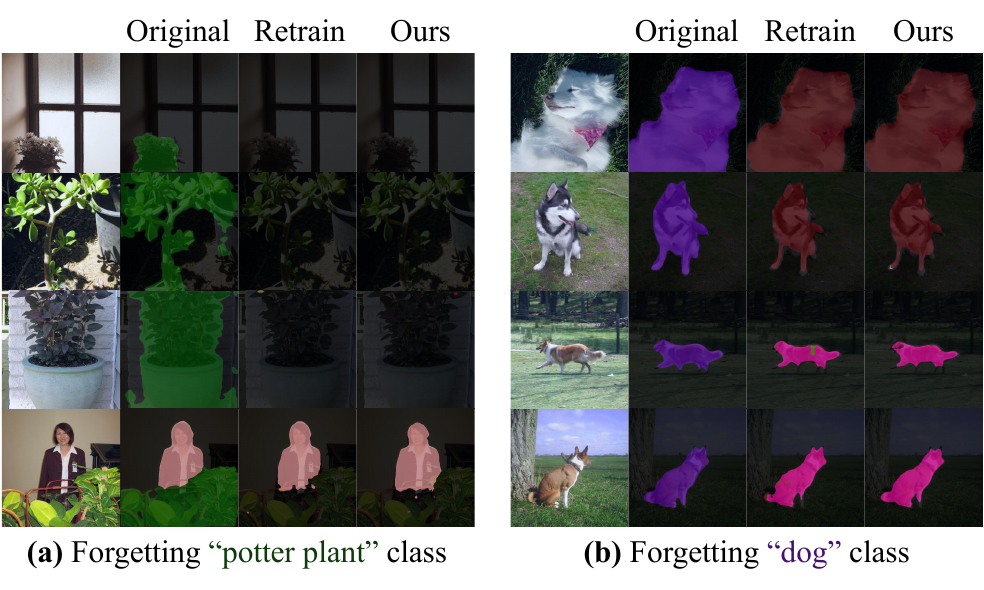}
    % \rule{0.8\columnwidth}{0.4\columnwidth} % 用 \rule 来创建一个占位框
    \vspace{-8mm}
    \caption{
    % Demonstration of machine unlearning in semantic segmentation.
    Visualization of unlearning in semantic segmentation.
    }
    \label{fig:seg}
    \vspace{-3mm}
\end{figure}

In this part, we apply machine unlearning to face recognition to achieve the objective of safeguarding individual privacy~\cite{vggface2, boundary}. 
As shown in the Appendix, our method outperforms all other methods across all metrics and achieves results closest to the retrain model in quantitative comparisons.
% Random label and boundary shrink also achieve great results.
% Although saliency unlearning achieve consistent results with ours in remaining accuracy, it resulted in a limited forgetting rate of only 11\% for \( \text{Acc}_{\mathrm{ft}} \).
To further investigate the forgetting effects, we employ Grad-CAM~\cite{gradcam} to visualize model attention areas. The heatmaps highlight areas important to each model’s predictions.
In {\cref{fig:gradcam}\smash{\textcolor{customblue}{a}}}, which relates to individuals designated for forgetting, the original model focuses on facial regions, while both the retrain model and our model effectively suppress recognition signals associated with these faces. In {\cref{fig:gradcam}\smash{\textcolor{customblue}{b}}}, depicting remaining individuals, all three models maintain strong attention on relevant features, indicating robust knowledge retention for non-forgotten classes.
%These Grad-CAM visualizations illustrate the effectiveness of our method in the face recognition task.

\vspace{-1mm}
\subsection{Backdoor Defense with Unlearning}
\vspace{-1mm}
Data poisoning is a common backdoor attack method.
Employing unlearning methods allows us to eliminate the influence of poisoned data and achieve effective defense.
% To evaluate the effectiveness of backdoor defense methods, we utilize two metrics: \textbf{ASR} (Attack Success Rate) and \textbf{Acc}. An effective defense algorithm should minimize ASR while maintaining a high Acc.
We adopt the recovery algorithm~\cite{backdoor} and utilize DELETE to unlearn.
% As shown in \cref{tab:backdoor defense}, our method achieves improved results. 
% as quanatative results in appendix, our method achieves improved results. 
The quantitative results in the Appendix show that DELETE achieves improved performance.
%Across various trigger sizes, our method shows performance gains ranging from 1\% to 4\% in most cases, with only a negligible difference in one instance. 
Notably, for 3×3 size, we observe improvements of 6.96\% and 4.19\% compared to others. 
Meanwhile, other methods face challenges of insufficient defense and reduced model performance.
% For example, NAD causes Acc to drop significantly to 69\% with a 3×3 trigger, while fine pruning yields an unsatisfactory defense effectiveness, with an ASR of 32.07\%.

\vspace{-1mm}
\subsection{Semantic Segmentation with Unlearning}
\vspace{-1mm}
% \begin{figure}[t]
%     \centering
%     \includegraphics[width=\columnwidth]{pic/exp/seg.pdf}
%     % \rule{0.8\columnwidth}{0.4\columnwidth} % 用 \rule 来创建一个占位框
%     \vspace{-8mm}
%     \caption{
%     % Demonstration of machine unlearning in semantic segmentation.
%     Visualization of unlearning in semantic segmentation.
%     }
%     \label{fig:seg}
%     \vspace{-4mm}
% \end{figure}

%We further investigate the application of machine unlearning in the challenging task of semantic segmentation. 
We compare the segmented output images of the original model, the retrain model, and ours, as illustrated in \cref{fig:seg}. Our model achieves segmentation results similar to those of the retrain model, effectively avoiding the segmentation of the target class.
Notably, forgetting a class is not the same as merely mis-segmenting it as background. For example, due to the absence of dog training samples, the retrained model may classify \textcolor{dog}{dog} as other visually similar categories, such as \textcolor{cat}{cat} or \textcolor{cow}{cow} in {\cref{fig:seg}\smash{\textcolor{customblue}{b}}}. Our approach effectively reproduces this behavior.
The quantitative comparison results are provided in the Appendix.
%Moreover, our method achieves an IoU of 0\% for the forget class and 68.93\% for the remain class, in a range similar to the retrained model's 0\% and 71.55\%, respectively.

% \begin{table}[t]
%     \centering
%     \caption{Comparison of Performance Metrics on Target and Remaining Classes}
%     \resizebox{0.9\columnwidth}{!}{
%         \begin{tabular}{l|ccc|ccc}
%             \toprule
%              \multirow{2}{*}{Method} & \multicolumn{3}{c|}{Target Class} & \multicolumn{3}{c}{Remaining Class} \\
%             \cmidrule(lr){2-4} \cmidrule(lr){5-7}
%              & Acc & IoU & fwIoU & Acc & mIoU & fwIoU \\
%             \midrule
%             \multicolumn{7}{c}{\textbf{forget xxx}} \\
%             \midrule
%             Retrain & - & - & - & - & - & - \\
%             Ours    & - & - & - & - & - & - \\
%             \midrule
%             \multicolumn{7}{c}{\textbf{forget xxx}} \\
%             \midrule
%             Retrain & - & - & - & - & - & - \\
%             Ours    & - & - & - & - & - & - \\
%             \midrule
%             \multicolumn{7}{c}{\textbf{forget xxx}} \\
%             \midrule
%             Retrain & - & - & - & - & - & - \\
%             Ours    & - & - & - & - & - & - \\
%         \end{tabular}
%     }
%     \label{tab:target_remain_metrics}
% \end{table}

\vspace{-2mm}
\section{Conclusion}
\vspace{-1mm}
\label{sec:conclusion}
In this paper, we introduce a novel machine unlearning method DELETE that requires neither access to remaining data nor intervention during training, aligning closely with real-world application constraints. 
Leveraging a theoretical framework that decomposes unlearning loss into forgetting and retention terms, our method effectively separates the influence of forgotten and retained classes through a masking distillation approach. Extensive experiments demonstrate its state-of-the-art performance across setups. 
We also apply our method to downstream tasks, including face recognition, backdoor unlearning, and semantic segmentation, demonstrating its real-world potential.

\section*{Acknowledgements}
This work was supported partially by NSFC(92470202, U21A20471), National Key Research and Development Program of China (2023YFA1008503), Guangdong NSF Project (No. 2023B1515040025).
% \sout{In this paper, we introduce a novel machine unlearning method that requires neither access to remaining data nor intervention during training, aligning closely with real-world application constraints. Leveraging a theoretical framework that decomposes unlearning loss into forgetting and retention terms, our method effectively separates the influence of forgotten and retained classes through a masking distillation approach. Extensive experiments demonstrate state-of-the-art performance across various datasets, model architectures, and both single-class and multi-class setups. We further apply our method to downstream tasks, including face recognition, backdoor unlearning, and semantic segmentation, highlighting its potential as a practical solution for real-world applications.
% }

% \newpage        % WARNING: delete this

{
    \small
    \bibliographystyle{ieeenat_fullname}
    \bibliography{main}

\begin{thebibliography}{72}
\providecommand{\natexlab}[1]{#1}
\providecommand{\url}[1]{\texttt{#1}}
\expandafter\ifx\csname urlstyle\endcsname\relax
  \providecommand{\doi}[1]{doi: #1}\else
  \providecommand{\doi}{doi: \begingroup \urlstyle{rm}\Url}\fi

\bibitem[Acquisti(2010)]{economics}
Alessandro Acquisti.
\newblock The economics of personal data and the economics of privacy.
\newblock \emph{Economics}, 2010.

\bibitem[Baumhauer et~al.(2022)Baumhauer, Sch{\"o}ttle, and Zeppelzauer]{baumhauer2022machine}
Thomas Baumhauer, Pascal Sch{\"o}ttle, and Matthias Zeppelzauer.
\newblock Machine unlearning: Linear filtration for logit-based classifiers.
\newblock \emph{Machine Learning}, 2022.

\bibitem[Bourtoule et~al.(2021)Bourtoule, Chandrasekaran, Choquette-Choo, Jia, Travers, Zhang, Lie, and Papernot]{sisa}
Lucas Bourtoule, Varun Chandrasekaran, Christopher~A Choquette-Choo, Hengrui Jia, Adelin Travers, Baiwu Zhang, David Lie, and Nicolas Papernot.
\newblock Machine unlearning.
\newblock In \emph{S\&P}, 2021.

\bibitem[Brophy and Lowd(2021)]{brophy2021machine}
Jonathan Brophy and Daniel Lowd.
\newblock Machine unlearning for random forests.
\newblock In \emph{ICML}, 2021.

\bibitem[Cao et~al.(2018)Cao, Shen, Xie, Parkhi, and Zisserman]{vggface2}
Qiong Cao, Li Shen, Weidi Xie, Omkar~M Parkhi, and Andrew Zisserman.
\newblock Vggface2: A dataset for recognising faces across pose and age.
\newblock In \emph{FG}, 2018.

\bibitem[Cao et~al.(2024)Cao, Liu, Zhang, Qiao, and Dong]{cao2024grids}
Shuo Cao, Yihao Liu, Wenlong Zhang, Yu Qiao, and Chao Dong.
\newblock Grids: Grouped multiple-degradation restoration with image degradation similarity.
\newblock In \emph{ECCV}, 2024.

\bibitem[Cao and Yang(2015)]{cao2015towards}
Yinzhi Cao and Junfeng Yang.
\newblock Towards making systems forget with machine unlearning.
\newblock In \emph{S\&P}, 2015.

\bibitem[Cha et~al.(2024)Cha, Cho, Hwang, Lee, Moon, and Lee]{learn2unlearn}
Sungmin Cha, Sungjun Cho, Dasol Hwang, Honglak Lee, Taesup Moon, and Moontae Lee.
\newblock Learning to unlearn: Instance-wise unlearning for pre-trained classifiers.
\newblock In \emph{AAAI}, 2024.

\bibitem[Chen et~al.(2018)Chen, Zhu, Papandreou, Schroff, and Adam]{deeplab}
Liang{-}Chieh Chen, Yukun Zhu, George Papandreou, Florian Schroff, and Hartwig Adam.
\newblock Encoder-decoder with atrous separable convolution for semantic image segmentation.
\newblock In \emph{ECCV}, 2018.

\bibitem[Chen et~al.(2023)Chen, Gao, Liu, Peng, and Wang]{boundary}
Min Chen, Weizhuo Gao, Gaoyang Liu, Kai Peng, and Chen Wang.
\newblock Boundary unlearning: Rapid forgetting of deep networks via shifting the decision boundary.
\newblock In \emph{CVPR}, 2023.

\bibitem[Chen et~al.(2019)Chen, Xiong, Xu, and Zuo]{chen2019novel}
Yuantao Chen, Jie Xiong, Weihong Xu, and Jingwen Zuo.
\newblock A novel online incremental and decremental learning algorithm based on variable support vector machine.
\newblock \emph{Cluster Computing}, 2019.

\bibitem[Chundawat et~al.(2023)Chundawat, Tarun, Mandal, and Kankanhalli]{badteacher}
Vikram~S Chundawat, Ayush~K Tarun, Murari Mandal, and Mohan Kankanhalli.
\newblock Can bad teaching induce forgetting? unlearning in deep networks using an incompetent teacher.
\newblock In \emph{AAAI}, 2023.

\bibitem[Crawford and Paglen(2021)]{ExcavatingAi}
Kate Crawford and Trevor Paglen.
\newblock Excavating {AI:} the politics of images in machine learning training sets.
\newblock \emph{{AI} Soc.}, 2021.

\bibitem[Dosovitskiy(2020)]{vit}
Alexey Dosovitskiy.
\newblock An image is worth 16x16 words: Transformers for image recognition at scale.
\newblock \emph{arXiv preprint arXiv:2010.11929}, 2020.

\bibitem[Everingham et~al.(2010)Everingham, Gool, Williams, Winn, and Zisserman]{pascalvoc}
Mark Everingham, Luc~Van Gool, Christopher K.~I. Williams, John~M. Winn, and Andrew Zisserman.
\newblock The pascal visual object classes {(VOC)} challenge.
\newblock \emph{IJCV}, 2010.

\bibitem[Fan et~al.(2023)Fan, Liu, Zhang, Wong, Wei, and Liu]{salun}
Chongyu Fan, Jiancheng Liu, Yihua Zhang, Eric Wong, Dennis Wei, and Sijia Liu.
\newblock Salun: Empowering machine unlearning via gradient-based weight saliency in both image classification and generation.
\newblock \emph{arXiv preprint arXiv:2310.12508}, 2023.

\bibitem[Foster et~al.(2024)Foster, Schoepf, and Brintrup]{ssd}
Jack Foster, Stefan Schoepf, and Alexandra Brintrup.
\newblock Fast machine unlearning without retraining through selective synaptic dampening.
\newblock In \emph{AAAI}, 2024.

\bibitem[Gandikota et~al.(2023{\natexlab{a}})Gandikota, Materzynska, Fiotto-Kaufman, and Bau]{erasing}
Rohit Gandikota, Joanna Materzynska, Jaden Fiotto-Kaufman, and David Bau.
\newblock Erasing concepts from diffusion models.
\newblock In \emph{ICCV}, 2023{\natexlab{a}}.

\bibitem[Gandikota et~al.(2023{\natexlab{b}})Gandikota, Materzynska, Fiotto-Kaufman, and Bau]{gandikota2023erasing}
Rohit Gandikota, Joanna Materzynska, Jaden Fiotto-Kaufman, and David Bau.
\newblock Erasing concepts from diffusion models.
\newblock In \emph{CVPR}, 2023{\natexlab{b}}.

\bibitem[Ginart et~al.(2019)Ginart, Guan, Valiant, and Zou]{ginart2019making}
Antonio Ginart, Melody Guan, Gregory Valiant, and James~Y Zou.
\newblock Making ai forget you: Data deletion in machine learning.
\newblock In \emph{NeurIPS}, 2019.

\bibitem[Golatkar et~al.(2020)Golatkar, Achille, and Soatto]{fisherforget}
Aditya Golatkar, Alessandro Achille, and Stefano Soatto.
\newblock Eternal sunshine of the spotless net: Selective forgetting in deep networks.
\newblock In \emph{CVPR}, 2020.

\bibitem[Gou et~al.(2021)Gou, Yu, Maybank, and Tao]{kdSurvey}
Jianping Gou, Baosheng Yu, Stephen~J Maybank, and Dacheng Tao.
\newblock Knowledge distillation: A survey.
\newblock \emph{IJCV}, 2021.

\bibitem[Graves et~al.(2021)Graves, Nagisetty, and Ganesh]{amnesiac}
Laura Graves, Vineel Nagisetty, and Vijay Ganesh.
\newblock Amnesiac machine learning.
\newblock In \emph{AAAI}, 2021.

\bibitem[Gu et~al.(2017)Gu, Dolan{-}Gavitt, and Garg]{badnet}
Tianyu Gu, Brendan Dolan{-}Gavitt, and Siddharth Garg.
\newblock Badnets: Identifying vulnerabilities in the machine learning model supply chain.
\newblock \emph{CoRR}, 2017.

\bibitem[Hanzlik et~al.(2021)Hanzlik, Zhang, Grosse, Salem, Augustin, Backes, and Fritz]{maas1}
Lucjan Hanzlik, Yang Zhang, Kathrin Grosse, Ahmed Salem, Maximilian Augustin, Michael Backes, and Mario Fritz.
\newblock Mlcapsule: Guarded offline deployment of machine learning as a service.
\newblock In \emph{CVPR}, 2021.

\bibitem[He et~al.(2016)He, Zhang, Ren, and Sun]{resnet}
Kaiming He, Xiangyu Zhang, Shaoqing Ren, and Jian Sun.
\newblock Deep residual learning for image recognition.
\newblock In \emph{CVPR}, 2016.

\bibitem[Hinton(2015)]{kd}
Geoffrey Hinton.
\newblock Distilling the knowledge in a neural network.
\newblock \emph{arXiv preprint arXiv:1503.02531}, 2015.

\bibitem[Izzo et~al.(2021)Izzo, Smart, Chaudhuri, and Zou]{influenceunlearn}
Zachary Izzo, Mary~Anne Smart, Kamalika Chaudhuri, and James Zou.
\newblock Approximate data deletion from machine learning models.
\newblock In \emph{AISTATS}, 2021.

\bibitem[Krizhevsky et~al.(2009)Krizhevsky, Hinton, et~al.]{cifar}
Alex Krizhevsky, Geoffrey Hinton, et~al.
\newblock Learning multiple layers of features from tiny images.
\newblock \emph{Technical report}, 2009.

\bibitem[Kurmanji et~al.(2024)Kurmanji, Triantafillou, Hayes, and Triantafillou]{scrub}
Meghdad Kurmanji, Peter Triantafillou, Jamie Hayes, and Eleni Triantafillou.
\newblock Towards unbounded machine unlearning.
\newblock In \emph{NeurIPS}, 2024.

\bibitem[Le and Yang(2015)]{tinyimagenet}
Ya Le and Xuan Yang.
\newblock Tiny imagenet visual recognition challenge.
\newblock \emph{CS 231N}, 2015.

\bibitem[Li et~al.(2024)Li, Zhou, Gao, Chen, Fu, Zhang, and Shui]{survey3}
Na Li, Chunyi Zhou, Yansong Gao, Hui Chen, Anmin Fu, Zhi Zhang, and Yu Shui.
\newblock Machine unlearning: Taxonomy, metrics, applications, challenges, and prospects.
\newblock \emph{arXiv preprint arXiv:2403.08254}, 2024.

\bibitem[Li et~al.(2021)Li, Lyu, Koren, Lyu, Li, and Ma]{nad}
Yige Li, Xixiang Lyu, Nodens Koren, Lingjuan Lyu, Bo Li, and Xingjun Ma.
\newblock Neural attention distillation: Erasing backdoor triggers from deep neural networks.
\newblock In \emph{ICLR}, 2021.

\bibitem[Liu et~al.(2018)Liu, Dolan{-}Gavitt, and Garg]{fineprune}
Kang Liu, Brendan Dolan{-}Gavitt, and Siddharth Garg.
\newblock Fine-pruning: Defending against backdooring attacks on deep neural networks.
\newblock In \emph{RAID}, 2018.

\bibitem[Liu et~al.(2022)Liu, Fan, Chen, Liu, Ma, Wang, and Ma]{backdoor}
Yang Liu, Mingyuan Fan, Cen Chen, Ximeng Liu, Zhuo Ma, Li Wang, and Jianfeng Ma.
\newblock Backdoor defense with machine unlearning.
\newblock In \emph{INFOCOM}, 2022.

\bibitem[Liu et~al.(2021)Liu, Lin, Cao, Hu, Wei, Zhang, Lin, and Guo]{swin-t}
Ze Liu, Yutong Lin, Yue Cao, Han Hu, Yixuan Wei, Zheng Zhang, Stephen Lin, and Baining Guo.
\newblock Swin transformer: Hierarchical vision transformer using shifted windows.
\newblock In \emph{ICCV}, 2021.

\bibitem[Liu et~al.(2023)Liu, Wu, Zheng, Lin, and Zheng]{liu2023generating}
Zuhao Liu, Xiao-Ming Wu, Dian Zheng, Kun-Yu Lin, and Wei-Shi Zheng.
\newblock Generating anomalies for video anomaly detection with prompt-based feature mapping.
\newblock In \emph{CVPR}, 2023.

\bibitem[Lv et~al.(2024)Lv, Long, Huang, Li, Lv, Ren, and Zheng]{lv2024spatialdreamer}
Zhen Lv, Yangqi Long, Congzhentao Huang, Cao Li, Chengfei Lv, Hao Ren, and Dian Zheng.
\newblock Spatialdreamer: Self-supervised stereo video synthesis from monocular input.
\newblock \emph{arXiv preprint arXiv:2411.11934}, 2024.

\bibitem[Mahadevan and Mathioudakis(2021)]{mahadevan2021certifiable}
Ananth Mahadevan and Michael Mathioudakis.
\newblock Certifiable machine unlearning for linear models.
\newblock \emph{arXiv preprint arXiv:2106.15093}, 2021.

\bibitem[Mo et~al.(2024)Mo, Gao, Fu, Yan, Wu, and Zheng]{mo2024bridge}
Qijie Mo, Yipeng Gao, Shenghao Fu, Junkai Yan, Ancong Wu, and Wei-Shi Zheng.
\newblock Bridge past and future: Overcoming information asymmetry in incremental object detection.
\newblock In \emph{ECCV}, 2024.

\bibitem[Nguyen et~al.(2022)Nguyen, Huynh, Nguyen, Liew, Yin, and Nguyen]{survey}
Thanh~Tam Nguyen, Thanh~Trung Huynh, Phi~Le Nguyen, Alan Wee-Chung Liew, Hongzhi Yin, and Quoc Viet~Hung Nguyen.
\newblock A survey of machine unlearning.
\newblock \emph{arXiv preprint arXiv:2209.02299}, 2022.

\bibitem[Oesterling et~al.(2024)Oesterling, Ma, Calmon, and Lakkaraju]{oesterling2024fair}
Alex Oesterling, Jiaqi Ma, Flavio Calmon, and Himabindu Lakkaraju.
\newblock Fair machine unlearning: Data removal while mitigating disparities.
\newblock In \emph{AISTATS}, 2024.

\bibitem[Qiao et~al.(2019)Qiao, Yang, and Li]{finetune}
Ximing Qiao, Yukun Yang, and Hai Li.
\newblock Defending neural backdoors via generative distribution modeling.
\newblock In \emph{NeurIPS}, 2019.

\bibitem[Romero et~al.(2014)Romero, Ballas, Kahou, Chassang, Gatta, and Bengio]{fitnet}
Adriana Romero, Nicolas Ballas, Samira~Ebrahimi Kahou, Antoine Chassang, Carlo Gatta, and Yoshua Bengio.
\newblock Fitnets: Hints for thin deep nets.
\newblock \emph{arXiv preprint arXiv:1412.6550}, 2014.

\bibitem[Salimans and Ho(2022)]{kdIndiffusion}
Tim Salimans and Jonathan Ho.
\newblock Progressive distillation for fast sampling of diffusion models.
\newblock \emph{arXiv preprint arXiv:2202.00512}, 2022.

\bibitem[Sanh(2019)]{kdInBert}
V Sanh.
\newblock Distilbert, a distilled version of bert: smaller, faster, cheaper and lighter.
\newblock \emph{arXiv preprint arXiv:1910.01108}, 2019.

\bibitem[Schramowski et~al.(2023)Schramowski, Brack, Deiseroth, and Kersting]{sld}
Patrick Schramowski, Manuel Brack, Bj{\"o}rn Deiseroth, and Kristian Kersting.
\newblock Safe latent diffusion: Mitigating inappropriate degeneration in diffusion models.
\newblock In \emph{CVPR}, 2023.

\bibitem[Schuhmann et~al.(2022)Schuhmann, Beaumont, Vencu, Gordon, Wightman, Cherti, Coombes, Katta, Mullis, Wortsman, et~al.]{laion}
Christoph Schuhmann, Romain Beaumont, Richard Vencu, Cade Gordon, Ross Wightman, Mehdi Cherti, Theo Coombes, Aarush Katta, Clayton Mullis, Mitchell Wortsman, et~al.
\newblock Laion-5b: An open large-scale dataset for training next generation image-text models.
\newblock In \emph{NeurIPS}, 2022.

\bibitem[Selvaraju et~al.(2017)Selvaraju, Cogswell, Das, Vedantam, Parikh, and Batra]{gradcam}
Ramprasaath~R Selvaraju, Michael Cogswell, Abhishek Das, Ramakrishna Vedantam, Devi Parikh, and Dhruv Batra.
\newblock Grad-cam: Visual explanations from deep networks via gradient-based localization.
\newblock In \emph{ICCV}, 2017.

\bibitem[Shan et~al.(2023)Shan, Cryan, Wenger, Zheng, Hanocka, and Zhao]{glaze}
Shawn Shan, Jenna Cryan, Emily Wenger, Haitao Zheng, Rana Hanocka, and Ben~Y Zhao.
\newblock Glaze: Protecting artists from style mimicry by text-to-image models.
\newblock \emph{arXiv preprint arXiv:2302.04222}, 2023.

\bibitem[Shen et~al.(2024)Shen, Zhang, Zhao, Bialkowski, Chen, and Xu]{laf}
Shaofei Shen, Chenhao Zhang, Yawen Zhao, Alina Bialkowski, Weitong~Tony Chen, and Miao Xu.
\newblock Label-agnostic forgetting: A supervision-free unlearning in deep models.
\newblock \emph{arXiv preprint arXiv:2404.00506}, 2024.

\bibitem[Shokri et~al.(2017)Shokri, Stronati, Song, and Shmatikov]{mia}
Reza Shokri, Marco Stronati, Congzheng Song, and Vitaly Shmatikov.
\newblock Membership inference attacks against machine learning models.
\newblock In \emph{S\&P}, 2017.

\bibitem[Simonyan and Zisserman(2014)]{vggnet}
Karen Simonyan and Andrew Zisserman.
\newblock Very deep convolutional networks for large-scale image recognition.
\newblock \emph{arXiv preprint arXiv:1409.1556}, 2014.

\bibitem[Song and Mittal(2021)]{miaasr}
Liwei Song and Prateek Mittal.
\newblock Systematic evaluation of privacy risks of machine learning models.
\newblock In \emph{USENIX}, 2021.

\bibitem[Tang et~al.(2024)Tang, Peng, Meng, and Zheng]{tang2024rethinking}
Yu-Ming Tang, Yi-Xing Peng, Jingke Meng, and Wei-Shi Zheng.
\newblock Rethinking few-shot class-incremental learning: Learning from yourself.
\newblock In \emph{ECCV}, 2024.

\bibitem[Tarun et~al.(2023)Tarun, Chundawat, Mandal, and Kankanhalli]{unsir}
Ayush~K Tarun, Vikram~S Chundawat, Murari Mandal, and Mohan Kankanhalli.
\newblock Fast yet effective machine unlearning.
\newblock \emph{TNNLS}, 2023.

\bibitem[Thudi et~al.(2022{\natexlab{a}})Thudi, Deza, Chandrasekaran, and Papernot]{unrollingsgd}
Anvith Thudi, Gabriel Deza, Varun Chandrasekaran, and Nicolas Papernot.
\newblock Unrolling sgd: Understanding factors influencing machine unlearning.
\newblock In \emph{EuroS\&P}, 2022{\natexlab{a}}.

\bibitem[Thudi et~al.(2022{\natexlab{b}})Thudi, Jia, Shumailov, and Papernot]{necessity}
Anvith Thudi, Hengrui Jia, Ilia Shumailov, and Nicolas Papernot.
\newblock On the necessity of auditable algorithmic definitions for machine unlearning.
\newblock In \emph{USENIX Security}, 2022{\natexlab{b}}.

\bibitem[Van~der Maaten and Hinton(2008)]{tsne}
Laurens Van~der Maaten and Geoffrey Hinton.
\newblock Visualizing data using t-sne.
\newblock \emph{JMLR}, 2008.

\bibitem[Voigt and Von~dem Bussche(2017)]{gdpr}
Paul Voigt and Axel Von~dem Bussche.
\newblock The eu general data protection regulation (gdpr).
\newblock \emph{A Practical Guide, 1st Ed., Cham: Springer International Publishing}, 2017.

\bibitem[Wang et~al.(2019)Wang, Yuan, Zhang, and Feng]{kdInDetect}
Tao Wang, Li Yuan, Xiaopeng Zhang, and Jiashi Feng.
\newblock Distilling object detectors with fine-grained feature imitation.
\newblock In \emph{CVPR}, 2019.

\bibitem[Weng et~al.(2022)Weng, Xiao, Yu, Wang, Wang, He, Li, Zhang, Lin, and Ding]{maas2}
Qizhen Weng, Wencong Xiao, Yinghao Yu, Wei Wang, Cheng Wang, Jian He, Yong Li, Liping Zhang, Wei Lin, and Yu Ding.
\newblock Mlaas in the wild: Workload analysis and scheduling in large-scale heterogeneous gpu clusters.
\newblock In \emph{NSDI}, 2022.

\bibitem[Wu et~al.(2023)Wu, Zheng, Liu, and Zheng]{wu2023estimator}
Xiao-Ming Wu, Dian Zheng, Zuhao Liu, and Wei-Shi Zheng.
\newblock Estimator meets equilibrium perspective: A rectified straight through estimator for binary neural networks training.
\newblock In \emph{ICCV}, 2023.

\bibitem[Wu et~al.(2024)Wu, Cai, Jiang, Zheng, Wei, and Zheng]{wu2024economic}
Xiao-Ming Wu, Jia-Feng Cai, Jian-Jian Jiang, Dian Zheng, Yi-Lin Wei, and Wei-Shi Zheng.
\newblock An economic framework for 6-dof grasp detection.
\newblock In \emph{ECCV}, 2024.

\bibitem[Wu et~al.(2020)Wu, Dobriban, and Davidson]{deltagrad}
Yinjun Wu, Edgar Dobriban, and Susan Davidson.
\newblock Deltagrad: Rapid retraining of machine learning models.
\newblock In \emph{ICML}, 2020.

\bibitem[Xu et~al.(2024)Xu, Wei, Zheng, Wu, and Zheng]{xu2024dexterous}
Guo-Hao Xu, Yi-Lin Wei, Dian Zheng, Xiao-Ming Wu, and Wei-Shi Zheng.
\newblock Dexterous grasp transformer.
\newblock In \emph{CVPR}, 2024.

\bibitem[Xu et~al.(2023)Xu, Zhu, Zhang, Zhou, and Yu]{mu:survey}
Heng Xu, Tianqing Zhu, Lefeng Zhang, Wanlei Zhou, and Philip~S. Yu.
\newblock Machine unlearning: A survey.
\newblock \emph{ACM Comput. Surv.}, 2023.

\bibitem[Zhao et~al.(2022)Zhao, Cui, Song, Qiu, and Liang]{dkd}
Borui Zhao, Quan Cui, Renjie Song, Yiyu Qiu, and Jiajun Liang.
\newblock Decoupled knowledge distillation.
\newblock In \emph{CVPR}, 2022.

\bibitem[Zhao et~al.(2024)Zhao, Ni, Fan, Wang, Chen, Meng, and Zhang]{gslora}
Hongbo Zhao, Bolin Ni, Junsong Fan, Yuxi Wang, Yuntao Chen, Gaofeng Meng, and Zhaoxiang Zhang.
\newblock Continual forgetting for pre-trained vision models.
\newblock In \emph{CVPR}, 2024.

\bibitem[Zheng et~al.(2024)Zheng, Wu, Yang, Zhang, Hu, and Zheng]{zheng2024selective}
Dian Zheng, Xiao-Ming Wu, Shuzhou Yang, Jian Zhang, Jian-Fang Hu, and Wei-Shi Zheng.
\newblock Selective hourglass mapping for universal image restoration based on diffusion model.
\newblock In \emph{CVPR}, 2024.

\bibitem[Zheng et~al.(2025{\natexlab{a}})Zheng, Huang, Liu, Zou, He, Zhang, Zhang, He, Zheng, Qiao, et~al.]{zheng2025vbench}
Dian Zheng, Ziqi Huang, Hongbo Liu, Kai Zou, Yinan He, Fan Zhang, Yuanhan Zhang, Jingwen He, Wei-Shi Zheng, Yu Qiao, et~al.
\newblock Vbench-2.0: Advancing video generation benchmark suite for intrinsic faithfulness.
\newblock \emph{arXiv preprint arXiv:2503.21755}, 2025{\natexlab{a}}.

\bibitem[Zheng et~al.(2025{\natexlab{b}})Zheng, Wu, Liu, Meng, and Zheng]{zheng2025diffuvolume}
Dian Zheng, Xiao-Ming Wu, Zuhao Liu, Jingke Meng, and Wei-shi Zheng.
\newblock Diffuvolume: Diffusion model for volume based stereo matching.
\newblock \emph{IJCV}, 2025{\natexlab{b}}.

\end{thebibliography}
}

\clearpage
% \def\pz{{\phantom{0}}}
% \newcommand\algcomment[1]{\def\@algcomment{\footnotesize#1}}
% \newcommand{\notice}[1]{
%     \colorbox{yellow}{\textcolor{red}{\textbf{Notice:}} #1}
% }
% \newcommand{\argmin}[1]{\mathop{\arg\min}\limits_{#1}}
% \newcommand{\g}[1]{\textcolor{gray!50}{#1}} % 用来表格中的灰色元素
% \newcommand{\am}[1]{\textcolor{blue}{#1}}   % after modify

% \newlength\savewidth\newcommand\shline{\noalign{\global\savewidth\arrayrulewidth
%   \global\arrayrulewidth 1pt}\hline\noalign{\global\arrayrulewidth\savewidth}}

\maketitlesupplementary
\renewcommand{\thetable}{S\arabic{table}}
\renewcommand{\thefigure}{S\arabic{figure}}
\renewcommand{\theequation}{S\arabic{equation}}
\renewcommand\thesection{\Alph{section}}

\noindent This appendix provides supplementary materials that could not be included in the main paper due to space limitations. 
In \cref{sec:reformulation}, we present the reformulation of the loss function Eq.~{(\textcolor{customblue}{5})}. 
\cref{sec:details} provides details on the implementation, baselines, and evaluation metrics.
\cref{sec:results} provides additional performance comparisons of various methods across different datasets and models. Finally, \cref{sec:application} demonstrates the implementation details, quantitative results, and qualitative results of applying our method to downstream tasks, including face recognition, backdoor defense, and semantic segmentation.

\section{Reformulation}
\label{sec:reformulation}
In this section, we reformulate the loss function Eq.~{(\textcolor{customblue}{5})} in the main paper. 
We prove that, with an appropriately designed mask \( \mathrm{Mask}_u'(\cdot) \), the masking and softmax steps can be interchanged. After reordering, the resulting softmax vector satisfies the condition that the sum of its elements equals 1, thereby obviating the need for additional normalization.

We define the mask function as \( \mathrm{Mask}_u'(\mathbf{v}) = \mathbf{v} + \mathbf{m}^{u} \), where the vector \( \mathbf{m}^{u} \in \mathbb{R}^K \) is defined as:

\[
\mathbf{m}_i^{u} = 
\begin{cases}
0 & \text{if } i \neq u, \\
-\infty & \text{if } i = u.
\end{cases}
\]

Next, we prove that interchanging the mask and softmax steps with an appropriately designed mask function results in proportional outcomes, \ie,
\[
\mathrm{Mask}_u(\mathrm{Softmax}(\mathbf z)) \propto \mathrm{Softmax}(\mathrm{Mask}_u'(\mathbf z))
\]
This is beacuse, for \( i \neq u \), we have:
$$
   \begin{aligned}
    \mathrm{Mask}_u(\mathrm{Softmax}(\mathbf z))_i %&= \mathrm{Mask}_u\left( \frac{e^{z_i}}{\sum_i e^{z_i}} \right) \\
    &= \frac{e^{z_i}}{\sum_i e^{z_i}} \\
    &= \frac{e^{z_i}}{\sum_{i \neq u} e^{z_i}} \times \frac{\sum_{i \neq u} e^{z_i}}{\sum_{i} e^{z_i}}.
    \end{aligned} 
    % \label{equ:case1}
$$
Meanwhile,
\[
\begin{aligned}
\mathrm{Softmax}(\mathrm{Mask}_u'(\mathbf z))_i &= \frac{e^{z_i}}{e^{-\infty} + \sum_{i \neq u} e^{z_i}} \\
&= \frac{e^{z_i}}{\sum_{i \neq u} e^{z_i}}.
\end{aligned}
\]
For \( i = u \), both element yield 0:
$$
    \mathrm{Mask}_u(\mathrm{Softmax}(\mathbf z))_i =\mathrm{Softmax}(\mathrm{Mask}_u'(\mathbf z))_i=0
    % \label{equ:case2}
$$
Thus, for any \( i \), combining the two cases above,
we obtain the following relationship:
\[
\mathrm{Mask}_u(\mathrm{Softmax}(\mathbf{z}))_i = \mathrm{Softmax}(\mathrm{Mask}_u'(\mathbf{z}))_i \times \frac{\sum_{i \neq u} e^{z_i}}{\sum_{i} e^{z_i}}.
\]

Since the elements of both vectors are proportional by a factor, normalizing these vectors results in identical outputs. Given that the softmax operation ensures the output vector is already normalized (\ie, its elements sum is 1), we conclude:
% \vspace{-2mm}
\begin{equation}
    \mathrm{Normalize}(\mathrm{Mask}_u(\mathrm{Softmax}(\mathbf z))) = \mathrm{Softmax}(\mathrm{Mask}_u'(\mathbf z)).
    \label{equ:interchange}
\end{equation}

Thus, we can interchange the mask and softmax steps via \cref{equ:interchange}, thereby simplifying Eq.~{(\textcolor{customblue}{5})} in the main paper by removing the need for normalization. The final loss function is expressed as:
\[
\begin{aligned}
\mathcal{L} 
&= \mathrm{KL}\left( \mathrm{Normalize}(\mathrm{Mask}_u(\mathrm{Softmax}(\mathbf z))) \,\middle\|\, \mathbf q \right) \\
% &= \mathrm{KL}\left( \mathrm{Normalize}(\mathrm{Softmax}(\mathrm{Mask}_u'(z))) \,\middle\|\,\mathbf q \right) \\
&= \mathrm{KL}\left( \mathrm{Softmax}(\mathrm{Mask}_u'(\mathbf z)) \,\middle\|\,\mathbf q \right).
\end{aligned}
\]

\section{Experimental Details}
\label{sec:details}
\subsection{Implementation Details}
Our implementation is based on Python 3.8 and PyTorch 1.13. All experiments are conducted on a system equipped with NVIDIA RTX 4090 GPU and Intel Xeon Gold 6226R CPU.
For the image classification unlearning task, we first train the original model from scratch. The optimizer settings are detailed in \cref{tab:optimize}. The training configurations for different models and datasets during pretraining are provided in \cref{tab:pretrain}. The retrain model follows the same training configurations but is trained exclusively on $\mathcal{D}_{\textrm{r}}$, without access to $\mathcal{D}_{\textrm{f}}$.

\begin{table}[t!]
\caption{Optimization settings for image classification.}
    \centering\small
    \setlength{\tabcolsep}{10pt}
    \begin{tabular}{l|l}
    % Config & Value \\ \shline
    Config & Value \\ 
    Optimizer & SGD \\
    % Base Learning rate & 0.1 \\
    Weight Decay& 5e-4 \\
    Momentum & 0.9 \\
    Learning Rate Scheduler & Step LR Scheduler \\
    Learning Rate Step & 40 \\
    Learning Rate Gamma & 0.1 \\
    % Training Epochs & 150\\
    % batch size & 128 \\
    \end{tabular}
\label{tab:optimize}
\end{table}

\begin{table}[t]
    \centering
    \caption{Pre-training setups for different datasets and models.}
    \resizebox{0.9\columnwidth}{!}{%
        \begin{tabular}{c|ccc}
            \toprule
            \multirow{2}{*}{Different Datasets}  & {CIFAR-10} & {CIFAR-100} & {Tiny ImageNet}  \\
                        & \small \g{ResNet-18}   & \small \g{ResNet-18}     & \small \g{ResNet-18} \\
            \midrule
            Pretrain Epochs & 150 & 150 & 150 \\
            Pretrain LR & 0.1 & 0.1 & 0.1 \\
            Batch Size  & 128 & 128 & 64  \\
            \midrule
            \multirow{2}{*}{Different Models}    & {VGG-16}    & {Swin-T}  & {ViT-S}   \\
                        & \small \g{CIFAR-10}    & \small \g{CIFAR-10}        & \small \g{CIFAR-10}   \\
            \midrule
            Pretrain Epochs & 80    & 100   & 150   \\
            Pretrain LR & 0.01  & 0.01  & 0.1   \\
            Batch Size  & 128   & 64    & 128   \\
            \bottomrule
        \end{tabular}%
    }
    % \vspace{-1mm}
    \label{tab:pretrain}
\end{table}
\subsection{Baseline Details}
Recalling that we compare the unlearning performance of different methods, under the two imposed constraints: \textbf{(\romannumeral 1)} No access to the remaining data; \textbf{(\romannumeral 2)} No intervention during the pre-training phase.  Under these constraints, certain methods become infeasible. Finetune, Fisher Forget~\cite{fisherforget}, SSD~\cite{ssd}, and SISA~\cite{sisa} fail to work due to their reliance on remaining data or intervention in their algorithms. Bad Teacher~\cite{badteacher}, Saliency Unlearn\cite{salun}, SCRUB~\cite{scrub}, and UNSIR\cite{unsir} are unable to compute regularization loss or perform additional repair phases. 

Nevertheless, to better compare the performance of different methods, we select several well-known approaches and evaluate them under scenarios without the aforementioned constraints.
The selected methods include Finetune, Fisher Forget, Bad Teacher, and Saliency Unlearn, with their results marked in \g{gray} in the main paper. Specifically, for Bad Teacher and Saliency Unlearn, we also remove the regularization loss that depend on remaining data and test their performance under the imposed constraints. These results are marked in black.

For training-based unlearning methods, we perform unlearning for 20 epochs and search for the optimal learning rate within the range of $[10^{-7}, 10^{-2}]$. For parameter-scrubbing unlearning methods, we search for the hyperparameter $\alpha$ in the range of $[10^{-8}, 10^{-5}]$ for Fisher Forget, and in the range of $[10^{-2}, 10^{2}]$ for Influential Unlearn. All other hyperparameter settings follow the configurations in the responding original papers.

\subsection{Mertic Details}
Following prior work, we use membership inference attacks (MIA) success rate as a metric to evaluate the forgetting performance of the unlearn model~\cite{salun}. The MIA implementation is based on a prediction confidence-based attack method~\cite{miaasr}. To construct the dataset for training the MIA predictor, we sample a balanced binary classification dataset from $\mathcal{D}_{\textrm{r}}$ and $\mathcal{D}_{\textrm{rt}}$. The input consists of the confidence scores predicted by the model for the images, while the corresponding labels indicate whether each image originates from $\mathcal{D}_{\textrm{r}}$ or $\mathcal{D}_{\textrm{rt}}$.

During the evaluation phase, the confidence scores predicted by $f_{\theta}$ on $\mathcal{D}_{\textrm{f}}$ are used as input to the MIA predictor. The classification results of the MIA predictor are then used to assess how many samples in the forgetting dataset are still memorized by the model after unlearning. The MIA metric is defined as:
% 在性能测量阶段，我们将$f_{\theta}$在$\mathcal D_{\textrm{f}}$上预测的confidence作为MIA的输入。根据MIA的分类结果判断，遗忘后的模型中仍有多少样本被模型memorize。MIA的定义如下:
$$
\textrm{MIA} = \frac{\textrm{TP}}{|\mathcal{D}_{\textrm{f}}|}
$$
where it measures the proportion of samples in $\mathcal{D}_{\textrm{f}}$ that the MIA predictor classifies as still being memorized by the model. In other words, it represents the proportion of samples in $\mathcal{D}_{\textrm{f}}$ that were not successfully forgotten.

Note that our MIA evaluation is slightly different from Saliency Unlearn's~\cite{salun}. We measure the proportion of samples in $\mathcal{D}_{\textrm{f}}$ that have not been successfully forgotten, whereas Saliency Unlearn measures the proportion of successfully forgotten samples. We adopt this approach to maintain consistency with metrics such as $\textrm{Acc}_{\textrm{f}}$ and $\textrm{Acc}_{\textrm{ft}}$, where lower values indicate better forgetting performance.
% 用来表示所有待遗忘样本中有多少MIA评估为待遗忘样本，\ie,$\mathcal{D}_{\textrm{f}}$中被成功遗忘的样本比例。请注意我们MIA的测量和salun有所不同，我们测量 $\mathcal{D}_{\textrm{f}}$中成功遗忘的样本比例，salun中测量未能成功遗忘的样本比例。我们选择这种方式，为了和$\textrm{Acc}_{\textrm{f}}$、$\textrm{Acc}_{\textrm{ft}}$相一致，也就是越低表示遗忘性能越好。

\section{Experimental Results}
\label{sec:results}
\subsection{Performance on Different Datasets}
\begin{table}[t]
    \centering
    \caption{Performance comparison of single-class forgetting across different unlearning methods on Tiny ImageNet dataset. \textbf{bold} indicates the single best result among methods. The same notation applies hereafter.}
    \resizebox{\columnwidth}{!}{
        \begin{tabular}{cc|cccccc}
            \toprule
                \multicolumn{2}{c|}{Method}  
                & $\mathrm{Acc}_{\mathrm{f}} \downarrow$ 
                & $\mathrm{Acc}_{\mathrm{r}} \uparrow$ 
                & $\mathrm{Acc}_{\mathrm{ft}} \downarrow$ 
                & $\mathrm{Acc}_{\mathrm{rt}} \uparrow$ 
                & $\mathrm{H\text{-}Mean} \uparrow$ 
                & $\mathrm{MIA} \downarrow$ \\
            \midrule
            \multicolumn{2}{c|}{Original Model} & 100 & 99.98 & 66.0 & 64.15 & - & 99.4 \\
            \multicolumn{2}{c|}{Retrain Model} & 0 & 99.98 & 0 & 64.42 & 65.20 & 0 \\
            \midrule
            \multicolumn{2}{c|}{Random Label} & 3.2 & 97.58 & 2.0 & 58.84 & 61.31 & 0.2 \\
            \multicolumn{2}{c|}{Negative Gradient} & 7.8 & 95.36 & 2.0 & 56.24 & 59.87 & 3.0\\
            \multicolumn{2}{c|}{Boundary Shrink} & 5.2 & 97.49 & 2.0 & 57.85 & 60.77 & 0.6 \\
            \multicolumn{2}{c|}{Boundary Expand} & 8.6 & 98.89 & 2.0 & 58.88 & 61.33 & 0 \\
            \multicolumn{2}{c|}{Influence Unlearn} & 12.2 & 99.36 & 2 & 60.30 &62.09 & 1.6 \\
            \multicolumn{2}{c|}{Learn to Unlearn} & 1.0 & 85.57 & 0 & 50.73 & 57.37 & 0.2 \\
            % \multicolumn{2}{c|}{Fisher Forget} & - & - & - & - & - & - \\
            \multicolumn{2}{c|}{Bad Teacher} & 5.8 & 98.32 & 2.0 & 59.02 & 61.41 & 0.2 \\
            \multicolumn{2}{c|}{Saliency Unlearn} & 6.6 & 98.80 & 2.0 & 59.67 & 61.76 & 0.6 \\
            \midrule
            \multicolumn{2}{c|}{Ours} & \textbf{0.4} & \textbf{99.94} & 0 & \textbf{62.15} & \textbf{64.02} & 0 \\
            % \multicolumn{2}{c|}{Learn to Unlearn (IMP)} & - & - & - & - & - & - \\
            % \multicolumn{2}{c|}{SCRUB} & - & - & - & - & - & - \\
            % \multicolumn{2}{l|}{Finetune} & 7.8 & 95.36 & 2.0 & 56.24 & 59.87 & 3.0\\
            % \multicolumn{2}{l|}{Fisher Forget} & - & - & - & - & - & - \\
            \bottomrule
        \end{tabular}
    }
    \label{tab:tiny imagenet}
\end{table}

In the main paper, we present the results of single-class forgetting on CIFAR-10 and CIFAR-100. \cref{tab:tiny imagenet} further presents the performance on Tiny ImageNet across various methods for the single-class forgetting task. While most methods achieve some degree of forgetting, several exhibit poor retention of knowledge for the remaining classes. For example, Negative Gradient, Boundary Shrink, Boundary Expand, and Learn to Unlearn show a decline of around 5\% in $\mathrm{Acc}_{\mathrm{rt}}$ compared to the original model.

Notably, despite Learn to Unlearn performing well on CIFAR-10 and CIFAR-100, its performance drops significantly on Tiny ImageNet, falling from third place on CIFAR-10 to the lowest rank on Tiny ImageNet. This highlights its fragility and inconsistency across datasets.

In contrast, our method consistently demonstrates strong performance, achieving complete forgetting in $\mathrm{Acc}_{\mathrm{ft}}$ while maintaining excellent $\mathrm{Acc}_{\mathrm{rt}}$. Among all methods, ours is one of only two to surpass 60\% in $\mathrm{Acc}_{\mathrm{rt}}$, outperforming the second-best method by nearly 2\%. Our method outperforms all other approaches across all performance metrics on Tiny ImageNet. This demonstrates the superior balance of our approach between effective forgetting and retention of unrelated knowledge across various datasets.

\subsection{Performance on Different Models}
\begin{table}[t]
    \centering
    \caption{Performance comparison of single-class forgetting on VGG-16, across different unlearning methods on CIFAR-10.}
    % \vspace{-2mm}
    \resizebox{\columnwidth}{!}{
        \begin{tabular}{cc|cccccc}
            \toprule
            % \cmidrule(lr){3-8}
                \multicolumn{2}{c|}{\multirow{1}{*}{Method}}  
                & $\mathrm{Acc}_{\mathrm{f}} \downarrow$ 
                & $\mathrm{Acc}_{\mathrm{r}} \uparrow$ 
                & $\mathrm{Acc}_{\mathrm{ft}} \downarrow$ 
                & $\mathrm{Acc}_{\mathrm{rt}} \uparrow$ 
                & $\mathrm{H\text{-}Mean} \uparrow$ 
                & $\mathrm{MIA} \downarrow$ \\
            \midrule
            \multicolumn{2}{c|}{Original Model}     & 99.94 & 99.94 & 91.20 & 92.02 & -     & 99.88     \\
            \multicolumn{2}{c|}{Retrain Model}      & 0     & 99.71 & 0     & 92.49 & 91.84 & 0         \\
            \midrule
            \multicolumn{2}{c|}{Random Label}       & 0     & 96.35 & 0     & 88.16 & 89.65 & 0         \\
            \multicolumn{2}{c|}{Negative Gradient}  &\pz0.36& 92.11 &\pz0.20& 83.77 & 87.24 & \pz0.20   \\
            \multicolumn{2}{c|}{Boundary Shrink}    &\pz6.68& 91.15 &\pz6.80& 82.79 & 83.59 & \pz2.84   \\
            \multicolumn{2}{c|}{Boundary Expand}    &\pz2.82& 84.95 &\pz2.10& 77.27 & 82.76 & \pz1.38   \\
            \multicolumn{2}{c|}{Influence Unlearn}  &\pz1.86& 93.17 &\pz1.60& 84.74 & 87.10 & \pz1.30 \\
            \multicolumn{2}{c|}{Learn to Unlearn}   &\pz3.28& 94.10 &\pz3.00& 85.56 & 86.86 & \pz1.96   \\
            \multicolumn{2}{c|}{Bad Teacher}        &\pz7.14& 90.95 &\pz7.20& 82.90 & 83.45 & 0         \\
            \multicolumn{2}{c|}{Saliency Unlearn}   & 11.24 & 73.37 &10.90  & 69.09 & 74.27 & 0         \\
            \midrule
            \multicolumn{2}{c|}{Ours}               & 0     & \textbf{99.66} & 0     & \textbf{92.08} & \textbf{91.64} & 0         \\
            \bottomrule
        \end{tabular}
    }
    \label{tab:vggnet}
\end{table}
\begin{table}[t]
    \centering
    \caption{Performance comparison of single-class forgetting on Swin-T, across different unlearning methods on CIFAR-10.}
    % \vspace{-2mm}
    \resizebox{\columnwidth}{!}{
        \begin{tabular}{cc|cccccc}
            \toprule
            % \cmidrule(lr){3-8}
                \multicolumn{2}{c|}{\multirow{1}{*}{Method}}  
                & $\mathrm{Acc}_{\mathrm{f}} \downarrow$ 
                & $\mathrm{Acc}_{\mathrm{r}} \uparrow$ 
                & $\mathrm{Acc}_{\mathrm{ft}} \downarrow$ 
                & $\mathrm{Acc}_{\mathrm{rt}} \uparrow$ 
                & $\mathrm{H\text{-}Mean} \uparrow$ 
                & $\mathrm{MIA} \downarrow$ \\
            \midrule
            \multicolumn{2}{c|}{Original Model}     & 85.00 & 86.58 & 82.70 & 82.68 & -     & 82.94     \\
            \multicolumn{2}{c|}{Retrain Model}      & 0     & 86.41 & 0     & 82.77 & 82.73 & 0         \\
            \midrule
            \multicolumn{2}{c|}{Random Label}       &\pz3.68& 74.86 &\pz0.90& 70.42 & 75.68 & \pz4.68   \\
            \multicolumn{2}{c|}{Negative Gradient}  &\pz0.46& 74.60 &\pz0.10& 71.98 & 76.93 & \pz0.22   \\
            \multicolumn{2}{c|}{Boundary Shrink}    &\pz5.54& 62.94 &\pz2.50& 58.18 & 67.44 & 11.36     \\
            \multicolumn{2}{c|}{Boundary Expand}    &\pz2.58& 70.36 &\pz1.00& 65.67 & 72.81 & \pz3.26   \\
            \multicolumn{2}{c|}{Influence Unlearn}  &\pz0.04& 76.02 & 0     & 72.99 & 77.54 & 0         \\
            \multicolumn{2}{c|}{Learn to Unlearn}   &\pz0.52& 79.55 &\pz0.40& 76.50 & 79.29 & \pz0.14   \\
            \multicolumn{2}{c|}{Bad Teacher}        &\pz8.96& 74.77 &\pz6.30& 68.86 & 72.43 & \pz7.24   \\
            \multicolumn{2}{c|}{Saliency Unlearn}   & 11.16 & 76.88 &\pz8.00& 72.40 & 73.53 & 12.40     \\
            \midrule
            \multicolumn{2}{c|}{Ours}               & \textbf{0}  & \textbf{86.68} & 0     & \textbf{83.61} & \textbf{83.15} & 0         \\
            \bottomrule
        \end{tabular}
    }
    \label{tab:swin-t}
\end{table}
\begin{table}[t]
    \centering
    \caption{Performance comparison of single-class forgetting on ViT-S, across different unlearning methods on CIFAR-10.}
    % \vspace{-2mm}
    \resizebox{\columnwidth}{!}{
        \begin{tabular}{cc|cccccc}
            \toprule
            % \cmidrule(lr){3-8}
                \multicolumn{2}{c|}{\multirow{1}{*}{Method}}  
                & $\mathrm{Acc}_{\mathrm{f}} \downarrow$ 
                & $\mathrm{Acc}_{\mathrm{r}} \uparrow$ 
                & $\mathrm{Acc}_{\mathrm{ft}} \downarrow$ 
                & $\mathrm{Acc}_{\mathrm{rt}} \uparrow$ 
                & $\mathrm{H\text{-}Mean} \uparrow$ 
                & $\mathrm{MIA} \downarrow$ \\
            \midrule
            \multicolumn{2}{c|}{Original Model}     & 90.50 & 90.57 & 72.70 & 75.73 &   -   & 84.60     \\
            \multicolumn{2}{c|}{Retrain Model}      &   0   & 92.42 &   0   & 76.44 & 74.52 &   0       \\
            \midrule
            \multicolumn{2}{c|}{Random Label}       &\pz1.32& 77.38 &\pz1.80& 68.51 & 69.68 &\pz2.24    \\ 
            \multicolumn{2}{c|}{Negative Gradient}  &\pz0.38& 85.25 &\pz0.50& 73.34 & 72.77 &\pz0.34    \\
            \multicolumn{2}{c|}{Boundary Shrink}    &\pz1.84& 73.16 &\pz1.70& 65.59 & 68.19 &\pz1.66    \\
            \multicolumn{2}{c|}{Boundary Expand}    &\pz8.02& 75.86 &\pz7.80& 67.42 & 66.14 &\pz9.24    \\
            \multicolumn{2}{c|}{Influence Unlearn}  &\pz0.98& 87.09 &\pz0.90& 75.20 & 73.46 &\pz0.78    \\
            \multicolumn{2}{c|}{Learn to Unlearn}   &\pz0.46& 85.11 &\pz0.30& 73.20 & 72.80 &\pz0.36    \\
            \multicolumn{2}{c|}{Bad Teacher}        &\pz6.02& 74.66 &\pz5.40& 67.27 & 67.28 &\pz7.32    \\
            \multicolumn{2}{c|}{Saliency Unlearn}   &\pz6.22& 77.54 &\pz5.90& 68.46 & 67.62 &\pz7.14    \\
            \midrule
            \multicolumn{2}{c|}{Ours}               & \textbf{0}    & \textbf{90.74} & \textbf{0} & \textbf{77.21} & \textbf{74.89} & \textbf{0} \\
            \bottomrule
        \end{tabular}
    }
    \label{tab:vit-s}
\end{table}
This part compares the performance of various methods across different models. In \cref{tab:vggnet,tab:swin-t,tab:vit-s}, we present the forgetting performance of different methods on three distinct models, apart from ResNet-18 discussed in the main paper: VGG-16, Swin-T, and ViT-S.

Remarkably, our method achieves the best results across all 18 metrics on all three models. Among these, it achieves the sole best performance on 14 metrics. For the overall performance metric H\textrm{-}Mean, our method outperforms the second-best by margins of 1.99\%, 3.86\%, and 1.43\% on VGG-16, Swin-T, and ViT-S, respectively. Additionally, on $\textrm{Acc}_{\textrm{rt}}$, which measures the preservation of knowledge for unrelated classes, our method leads by 3.92\%, 7.11\%, and 1.99\% over the second-best on the respective models. These results highlight the strong generalization and consistent performance of our method across different architectures.

In contrast, some other methods exhibit significant performance fluctuations, reflecting their fragility when applied to different models. For example, Boundary Shrink shows a gap of 7.74\% in H\text{-}Mean compared to the retrain model on ResNet-18, which expands to 15.29\% on Swin-T, nearly doubling the performance gap. On the other hand, Influence Unlearn demonstrates relatively stable performance, achieving three second-place and one third-place rankings. However, its H\text{-}Mean scores are 1.22\% to 5.61\% lower than those of our method across all models, further emphasizing our superior performance.

\section{Application to Downstream Tasks}
\label{sec:application}
\subsection{Face Recognition with Unlearning}
\noindent\textbf{Implementation Details.}
Face recognition experiments are conducted using the VGGFace2~\cite{vggface2} dataset and the ResNet18 model. To prepare the dataset, we first filter individuals with more than 500 images, and randomly select 110 of them. For each individual, we allocate 400 images for the training set and 100 images for the test set.

The model is pretrained using 100 randomly selected individuals from the training set for 80 epochs to obtain a pretrained model. Subsequently, we fine-tune the model for 40 epochs using 10 other individuals. During the unlearning phase, one of these 10 individuals is randomly chosen, and the unlearning process is performed for 20 epochs. To ensure fairness, all methods are evaluated using the same individuals for the pretraining, fine-tuning, and unlearning stages.
The batch size is set to 64 for all training stages, and the learning rate is fixed at 0.01 during both the pretraining and fine-tuning phases.

% \noindent\textbf{Qualitative and Quantitative Results.}
\noindent\textbf{Qualitative Results.}
\begin{table}[t]
    \centering
    % \vspace{-3mm}
    \caption{Performance comparison of face recognition task on VGG-Face dataset.}
    \resizebox{\columnwidth}{!}{
        \begin{tabular}{cc|cccccc}
            \toprule
            % \cmidrule(lr){3-8}
                \multicolumn{2}{c|}{\multirow{1}{*}{Method}}  
                & $\mathrm{Acc}_{\mathrm{f}} \downarrow$ 
                & $\mathrm{Acc}_{\mathrm{r}} \uparrow$ 
                & $\mathrm{Acc}_{\mathrm{ft}} \downarrow$ 
                & $\mathrm{Acc}_{\mathrm{rt}} \uparrow$ 
                & $\mathrm{H\text{-}Mean} \uparrow$ 
                & $\mathrm{MIA} \downarrow$ \\
            \midrule
            \multicolumn{2}{c|}{Original Model} & 100 & 100 & 97.00 & 99.22 & - & 100 \\
            \multicolumn{2}{c|}{Retrain Model} & 0 & 100 & 0 & 98.89 & 97.94 & 0 \\
            \midrule
            \multicolumn{2}{c|}{Random Label} & 0 & 99.97 & 0 & 97.22 & 97.11 & 0 \\
            % \multicolumn{2}{c|}{Finetune} & 0 & 100 & 0 & 97.22 & 97.11 & 0 \\
            \multicolumn{2}{c|}{Negative Gradient} & \pz1.15 & 98.25 & \pz1.00 & 93.67 & 94.82 & \pz1.00 \\
            \multicolumn{2}{c|}{Boundary Shrink} & 0 & 100 & 0 & 97.33 & 97.16 & 0 \\
            \multicolumn{2}{c|}{Boundary Expand} & \pz4.25 & 99.94 & \pz3.00 & 97.33 & 95.64 & 0 \\
            \multicolumn{2}{c|}{Influence Unlearn} & 0 & 99.97 & 0 & 96.11 & 96.55 & 0 \\
            \multicolumn{2}{c|}{Learn to Unlearn} & \pz2.25 & 96.94 & \pz1.00 & 92.33 & 94.13 & \pz1.25 \\
            \multicolumn{2}{c|}{Bad Teacher} & \pz2.75 & 98.50 & \pz3.00 & 95.56 & 94.77 & 0 \\
            \multicolumn{2}{c|}{Saliency Unlearn} & 18.50 & 100 & 11.00 & 97.67 & 91.46 & \pz3.50 \\
            % \multicolumn{2}{c|}{Fisher Forget} & 48.00 & 48.41 & 37.00 & 46.44 & 52.36 & 80.25 \\
            \midrule
            \multicolumn{2}{c|}{Ours} & 0 & 100 & 0 & 97.67 & \textbf{97.33} & 0 \\
            \bottomrule
        \end{tabular}
    }
    \label{tab:vggface}
    % \vspace{-4mm}
\end{table}
\cref{tab:vggface} presents the performance of various methods on the face recognition task. Our method achieves the best results across all metrics. In terms of overall performance, measured by H\text{-}Mean, our method demonstrates a minimal gap of just 0.61\% compared to the retrain model, which is considered as the upper bound.

\subsection{Backdoor Defense with Unlearning}
\noindent\textbf{Implementation Details.}
Data poisoning is a common backdoor attack method, where a small fraction of the training data is maliciously altered with triggers and incorrect labels. Once trained on such poisoned data, the backdoored model behaves normally on clean inputs but misclassifies those containing the embedded trigger.

To defend against such attacks, unlearning methods aim to effectively remove the influence of poisoned data from the model. Following the setup~\cite{backdoor}, we adopt BadNets~\cite{badnet} as the backdoor attack scenario in our experiments. Specifically, we conduct experiments on the CIFAR-10 dataset using a ResNet18 model. In this setup, 5\% of the training data is poisoned by embedding triggers of specified sizes at random locations within the images.
This experiment allows us to evaluate the ability of unlearning methods to eliminate the backdoor effect while preserving the model’s performance on clean data.

In data poisoning scenarios, the trigger 
is often inaccessible. Consequently, our backdoor defense approach is divided into two phases:  recovering the trigger and unlearning it. We adopt the recovery algorithm from the BAERASER~\cite{backdoor} and replace the negative gradient method with our approach in the unlearning phase.

To evaluate the effectiveness of backdoor defense methods, we utilize two metrics: \textbf{ASR} and \textbf{Acc}. Here, ASR (Attack Success Rate) measures the success rate on the poisoned data, while Acc indicates the classification accuracy on clean samples. An effective defense algorithm should minimize ASR while maintaining a high Acc.

\noindent\textbf{Qualitative and Quantitative Results.}
As shown in \cref{tab:backdoor defense}, our method achieves improved Quantitative results. Across various trigger sizes, our method shows performance gains ranging from 3.00\% to 6.96\% on all metrics. Notably, for small 3×3 trigger sizes, we observe improvements of 6.96\% and 4.19\% compared to the recovery+NG method.
Meanwhile, some other methods face challenges of insufficient defense and reduced model performance.
For example, NAD causes Acc to drop significantly to 69\% with a 3×3 trigger, while fine pruning yields unsatisfactory defense effectiveness, with an ASR of 32.07\%.

\begin{figure}[t]
    \centering
    \includegraphics[width=0.9\columnwidth]{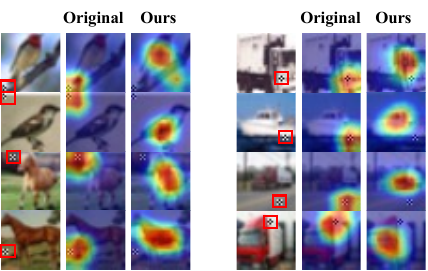}
    \caption{
    Visualization of attention heatmaps generated by the original backdoored model (\textbf{Original}) and the model repaired using our method (\textbf{Ours}). The red boxes
    {\protect\tikz[baseline] \protect\draw[red, very thick] (0,0) rectangle (0.2cm, 0.2cm);}
    highlight the backdoor triggers in the original images. The original backdoored model focuses primarily on the trigger regions, while our method successfully shifts attention to classification-relevant areas, demonstrating effective backdoor defense.
    \label{fig:backdoor}
    }
\end{figure}

We further present the attention visualizations of the backdoored model and the model repaired using our method. As shown in the \cref{fig:backdoor}, our method effectively erases the influence of poisoned data, enabling the repaired model to shift its focus away from the trigger and towards classification-relevant regions.

\begin{table}[t]
    \centering
    \caption{Attack success rate and accuracy across different trigger sizes and defense methods on CIFAR-10.}
    % \vspace{-2mm}
    \resizebox{\columnwidth}{!}{
        \begin{tabular}{c|cc|cc|cc}
            \toprule
            \multirow{2}{*}{Method} & \multicolumn{2}{c|}{{3$\times$3}} & \multicolumn{2}{c|}{{5$\times$5}} & \multicolumn{2}{c}{{7$\times$7}} \\
            \cmidrule(lr){2-3} \cmidrule(lr){4-5} \cmidrule(lr){6-7}
             & ASR $\downarrow$ & Acc $\uparrow$ & ASR $\downarrow$ & Acc $\uparrow$ & ASR $\downarrow$ & Acc $\uparrow$ \\
            \midrule
            Original Model  & 98.80 & 83.81 & 97.99 & 83.64 & 98.01 & 82.89 \\
            \midrule
            Fine Pruning~\cite{fineprune}   & 32.07 & 77.96 & 37.41 & 78.34 & 35.24 & 76.29 \\
            NAD~\cite{nad}                  & \pz4.24 & 69.00 & \pz7.68 & 69.25 & \pz8.95 & 70.49 \\
            Finetune~\cite{finetune}        & \pz9.52 & 78.95 & \pz9.32 & 78.51 & 10.77 & 78.32 \\
            Recovery + NG                   & \pz7.96 & 79.31 & \pz5.64 & 79.59 & \pz4.66 & {79.19} \\
            \midrule
            Recovery + Ours & \pz\textbf{1.00} & \textbf{83.50} & \pz\textbf{1.24} & \textbf{83.07} & \pz\textbf{1.66} & \textbf{82.23} \\
            \bottomrule
        \end{tabular}
    }
    \label{tab:backdoor defense}
    % \vspace{-5mm}
\end{table}

\subsection{Semantic Segmentation with Unlearning}
\noindent\textbf{Implementation Details.}
In this part, we explore the application of machine unlearning in the downstream task of semantic segmentation. Semantic segmentation aims to classify every pixel in an image, thereby generating a segmentation map. We use the DeepLab v3+ model~\cite{deeplab} and the PASCAL VOC dataset~\cite{pascalvoc} to conduct segmentation experiments.

Similar to unlearning in classification, we first train a segmentation model from scratch, and then unlearn an entire class from it. During the pretraining phase, we employ SGD optimization with a learning rate of 0.007, weight decay of 5e-4, and train for 50 epochs.
Then we perform unlearning over 10 epochs.
%In the unlearning phase, we perform unlearning for 10 epochs. %For different unlearning methods and forgotten classes, we search for the optimal learning rate within the range of $[10^{-4}, 10^{-2}]$.

To evaluate the methods' unlearning and remaining performance, we measure the Intersection over Union (IoU) on the test set. The IoU quantifies the overlap between the model's predicted segmentation map and the ground truth; a higher IoU indicates better segmentation performance. We compute the IoU for the forgotten class, denoted as {$\textbf{IoU}_{\textbf{ft}}$}, and the average IoU for the remaining classes, denoted as $\textbf{IoU}_{\textbf{rt}}$. In machine unlearning, a lower $\textrm{IoU}_{\textrm{ft}}$ indicates more effective unlearning, while a higher $\textrm{IoU}_{\textrm{rt}}$ suggests that the knowledge on remaining classes is less affected.

\noindent\textbf{Qualitative and Quantitative Results.}
In terms of quantitative results, as shown in \cref{tab:seg}, our method achieves 0\% IoU for the forgotten class and 68.93\% IoU for the remaining classes, delivering performance closest to that of the retrain model. Other methods, while failing to effectively unlearn, exhibit a significant performance drop of over 25\% in $\textrm{IoU}_{\textrm{rt}}$, indicating their inability to achieve effective unlearning or sufficient preservation of the remaining knowledge.

\begin{figure}[t]
    \centering
    \includegraphics[width=0.9\columnwidth]{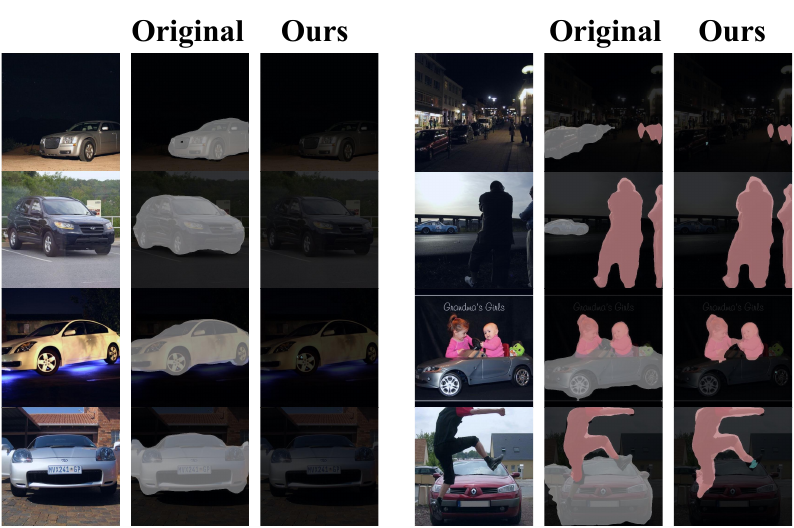}
    \caption{
    Visualization of machine unlearning in the semantic segmentation task, including segmentation results generated by the original model (\textbf{Original}) and the unlearned model with our method (\textbf{Ours}).
    }
    \label{fig:seg car}
\end{figure}

We further present the visualization of segmented results before and after unlearning in \cref{fig:seg car}. The left three columns show segmentation results for images containing only the \textcolor{gray!70}{car} class. While the original model correctly identifies and outlines the \textcolor{gray!70}{car}, the unlearned model has “forgotten” and no longer detects it. The right three columns depict images with \textcolor{gray!70}{car} and other objects, demonstrating that the target class is successfully forgotten, and the segmentation of \textcolor{customPink}{people} remains preserved.
\begin{table}[t]
    \centering
    \caption{Performance comparison of different unlearning methods on the semantic segmentation task.}
    \resizebox{\columnwidth}{!}{
        \begin{tabular}{l|c|ccc}
            \toprule
            Metric &  Retrain Model & Negative Gradient &  Random  Label &  Ours  \\
            \midrule
            $\textrm{IoU}_{\textrm{ft}}$ $\downarrow$  & 0 & 10.26 & 22.26 & \textbf{0} \\
            $\textrm{IoU}_{\textrm{rt}}$ $\uparrow$    & 71.55 & 42.26 & 45.06 & \textbf{68.93} \\
            \bottomrule
        \end{tabular}
    }
    \label{tab:seg}
    \vspace{-2mm}
\end{table}

\clearpage

\end{document}